\documentclass[final]{article}

\usepackage{microtype}
\usepackage{graphicx}
\usepackage{subfigure}
\usepackage{booktabs} 

\usepackage[accepted]{icml2023}

\usepackage{multirow}

\usepackage{amsmath}
\usepackage{amssymb}

\usepackage{mathtools}
\usepackage{amsthm}
\usepackage{graphicx}
\usepackage[capitalize,noabbrev]{cleveref}

\theoremstyle{plain}

\theoremstyle{definition}

\theoremstyle{remark}

\usepackage[textsize=tiny]{todonotes}

\icmltitlerunning{Blackout Diffusion: Generative Diffusion Models in Discrete-State Spaces}

\begin{document}

\twocolumn[
\icmltitle{Blackout Diffusion: Generative Diffusion Models in Discrete-State Spaces}




\begin{icmlauthorlist}
\icmlauthor{Javier E.~Santos}{EES16}
\icmlauthor{Zachary R.~Fox}{CCS3,ORNL}
\icmlauthor{Nicholas Lubbers}{CCS3}
\icmlauthor{Yen Ting Lin}{CCS3}

\end{icmlauthorlist}

\icmlaffiliation{EES16}{Computational Earth Science Group (EES-16), Earth and Environmental Sciences Division, Los Alamos National Laboratory, Los Alamos, NM 87545, USA}

\icmlaffiliation{CCS3}{Information Sciences Group (CCS-3), Computer, Computational and Statistical Sciences Division, Los Alamos National Laboratory, Los Alamos, NM 87545, USA}

\icmlaffiliation{ORNL}{\emph{Currently at} Oak Ridge National Laboratory, Oak Ridge, TN 37830, USA}


\icmlcorrespondingauthor{Yen Ting Lin}{yentingl@lanl.gov}

\icmlkeywords{Generative Diffusion Model, Discrete-state Markov Processes, Master Equations}

\vskip 0.3in
]



\printAffiliationsAndNotice{}  

\begin{abstract}
Typical generative diffusion models rely on a Gaussian diffusion process for training the backward transformations, which can then be used to generate samples from Gaussian noise. However, real world data often takes place in discrete-state spaces, including many scientific applications.  Here, we develop a theoretical formulation for arbitrary discrete-state Markov processes in the forward diffusion process using exact (as opposed to variational) analysis. We relate the theory to the existing continuous-state Gaussian diffusion as well as other approaches to discrete diffusion, and identify the corresponding reverse-time stochastic process and score function in the continuous-time setting, and the reverse-time mapping in the discrete-time setting. As an example of this framework, we introduce ``Blackout Diffusion'', which learns to produce samples from an empty image instead of from noise. Numerical experiments on the CIFAR-10, Binarized MNIST, and CelebA datasets confirm the feasibility of our approach. Generalizing from specific (Gaussian) forward processes to discrete-state processes without a variational approximation sheds light on how to interpret diffusion models, which we discuss.
\end{abstract}

\section{Introduction}
Diffusion processes have been recently utilized to construct \textit{Diffusion Models}, a class of generative models in deep learning \cite{sohl-dicksteinDeepUnsupervisedLearning2015,hoDenoisingDiffusionProbabilistic2020,songScoreBasedGenerativeModeling2021a}. These frameworks consist of a set of trainable transformations (implemented as deep neural networks) that sequentially process a prescribed distribution (the \emph{prior}, usually a high-dimensional isotropic Gaussian) to the data distribution. To train the networks, samples drawn from the data distribution are transformed by a stochastic process, which goes forward in time, and which has the prior as a stationary (final) distribution. Realizations from the forward diffusion process are used to train transformations that approximate the reverse-time process. In generative inference, samples are drawn from the prior, and the trained network is used to transform them into samples of the learned data distribution. 

Diffusion Models have been used for many applications, including image \cite{sohl-dicksteinDeepUnsupervisedLearning2015,hoDenoisingDiffusionProbabilistic2020,nicholImprovedDenoisingDiffusion2021,songGenerativeModelingEstimating2020,songImprovedTechniquesTraining2020,songScoreBasedGenerativeModeling2021a,songMaximumLikelihoodTraining2021}, audio \cite{kong2020diffwave}, video \cite{hoppe2022diffusion,ho2022video}, and language \cite{gong2022diffuseq}. 
Most works follow the original formulation \cite{sohl-dicksteinDeepUnsupervisedLearning2015,hoDenoisingDiffusionProbabilistic2020}, utilizing Gaussian diffusion on continuous domains. However, there are wide-ranging data domains which are not continuous in nature\footnote{Even digital images are usually encoded using quantized level-values for each pixel.}, and applying quantization and de-quantization for treating these data may not be ideal. For example, single-molecule and single-cell gene expressions \cite{munsky2015integrating,Pichon18} study systems with very small counts, in which discrete effects are qualitatively relevant to system behavior. As another example, phase-separated fluid problems occur in many applications pertaining to industrial and earth sciences, where each region of space is occupied categorically, that is, by exactly one of several fluid types \cite{kanglbm}. Graph structures are pervasive and machine learning on these structures is an area of tremendous recent growth \cite{graphs_MethodsApps}, for example being the dominant representation for molecular structure, which is of relevance to chemistry \cite{gilmer17}, drug discovery \cite{WIEDER20201, drugdiscoveryai}, and biophysics \cite{jiang2021interactiongraphnet}. In bioinformatics, both DNA and protein sequences are codes consisting of discrete values \cite{ingraham2019generative}. 

Some prior works have investigated diffusion modeling beyond the Gaussian paradigm. For example, \citet{Bansal2022ColdNoise} challenged the notion that noise is required by showing that deterministic degradation (e.g. blurring) applied to images, constructing a deterministic alternative, and discarding the stochastic theory entirely. Despite the abundance of potential applications, few theoretical formulations diffusion modeling of arbitrary discrete-state systems have been proposed. \citet{sohl-dicksteinDeepUnsupervisedLearning2015} explores a particular form of binomial diffusion kernels as a discrete-time diffusion. 
 \citet{Ye2022} proposed leveraging the exit distribution of first-passage processes to achieve discrete-state generative modeling. Termed as the First Hitting Diffusion Models (FHDM), the underlying process is still a continuous-state Gaussian (It\^o) diffusion, but FHDM can sample discrete-state distributions without ad hoc quantization and dequantization. Most relevant to our work include (1) Multinomial Diffusion \cite{hoogeboom2021argmax}, which invoked a specific discrete Markovian process (the multinomial process) and solved for the corresponding reverse process analytically, (2) \citet{austin21} created Masked Diffusion using a discrete-state Markov chain as the forward diffusion which transforms samples to a unique ``masked state'', and (3) the recent work by \citet{campbell2022} described a theoretical framework for continuous-time discrete-state Generative Diffusion Modeling. Importantly, the theories developed in both \citet{austin21,campbell2022} take a variational inference approach to constructing the loss function, and \citet{campbell2022} achieves a continuous-time formulation by taking a limit of this approach. An exact (i.e. non-variational) analysis has not been seen in the existing literature.
 
Addressing diffusion for arbitrary discrete-state processes requires answering several questions. For example, is there a discrete-state formula corresponding to the Brownian bridge\footnote{\citet{hoDenoisingDiffusionProbabilistic2020} used a discrete-time formulation and Bayes formula to derive the reverse map. The same technique is often used for constructing conditional distributions for It\^o processes, and the derived formula is commonly known as the Brownian bridge\cite{revuzContinuousMartingalesBrownian1994} because it connects initial and final conditions, resulting in an intermediate-time (`bridge') Gaussian distribution, although this connection to the technique was not pointed out in \citet{hoDenoisingDiffusionProbabilistic2020}.} used in \citet{hoDenoisingDiffusionProbabilistic2020} for learning the reverse-time mappings, and more broadly, is diffusion modeling only possible when there are closed-form solutions for reverse-time mapping? What is the Stein score function \cite{songGenerativeModelingEstimating2020,songImprovedTechniquesTraining2020,songScoreBasedGenerativeModeling2021a,songMaximumLikelihoodTraining2021} for learning the reverse-time process? Our contribution is to answer these questions and put forth a general and \textit{exact} theoretical framework for constructing Diffusion Models in discrete-state spaces for an arbitrary Markov process, which can be either discrete-time or continuous-time in nature. Our core contribution is a prescription for the generator of the reverse-time stochastic process, forming the discrete analog to \citet{andersonReversetimeDiffusionEquation1982} for It\^o Stochastic Differential Equations (SDEs).

We then examine explicitly a specific case, the \textit{pure-death} process, which corresponds physically to radioactive decay. In this case, the prior consists of a single point---a completely black image, similar to Mask Diffusion \cite{austin21}.  Based on this characteristic, we call this approach \textit{Blackout Diffusion}. We emphasize that our process is different from Mask Diffusion \cite{austin21}, although both processes (almost-surely) converge to a singular state as $t\rightarrow \infty$. We show that Blackout Diffusion can learn and generate the CIFAR-10, CelebA, and Binarized MNIST datasets without interpreting the data as continuous at any stage. Additionally, we consider how discrete-state models answer two additional conceptual questions regarding generative Diffusion Models: (1) Do the forward and reverse stochastic processes have to correspond to noisifying and denoising processes? (2) Is it natural to consider the prescribed prior as a latent-space representation, as can be done for normalizing flows \cite{rezende2015variational} and variational autoencoders \cite{kingmaAutoencodingVariationalBayes2014}?

The rest of the paper is structured as follows. Section~\ref{sec:method} presents the construction of discrete-space forward, backward and reverse processes, loss functions and score functions, and the relationship to Gaussian diffusion. For the rest of the article, we choose a particular Markov model, Blackout Diffusion defined in Sec.~\ref{sec:pureDeathProcess}, presenting results in Sec.~\ref{sec:numerics} and questions raised in Sec.~\ref{sec:discussion}.

\section{Method} \label{sec:method}

There are several key components to tractable Diffusion Models. First, one must be able to sample the forward stochastic process efficiently for generating the training data. Secondly, one needs to analytically prescribe summary statistics for training the backwards transformations; learning from individual samples would be computationally prohibitive. Existing generative Diffusion Models \cite{sohl-dicksteinDeepUnsupervisedLearning2015,hoDenoisingDiffusionProbabilistic2020,nicholImprovedDenoisingDiffusion2021,songGenerativeModelingEstimating2020,songImprovedTechniquesTraining2020,songScoreBasedGenerativeModeling2021a,songMaximumLikelihoodTraining2021} use a particular Gaussian diffusion process which is applied independently in the dimension of the data, e.g.~in each color channel of each pixel in an image, for which the theoretical formulation is complete. For discrete-time implementations \cite{sohl-dicksteinDeepUnsupervisedLearning2015,nicholImprovedDenoisingDiffusion2021}, the forward solution is analytically tractable, and the summary statistics---the conditional mean and variance of the reverse-time mapping---can be obtained via the Brownian bridge technique \cite{revuzContinuousMartingalesBrownian1994}. For continuous-time implementations \cite{songGenerativeModelingEstimating2020,songImprovedTechniquesTraining2020,songScoreBasedGenerativeModeling2021a,songMaximumLikelihoodTraining2021}, the forward solution is also tractable, and the summary statistics (the drift and diffusion of SDE) for the reverse-time process were provided by \citet{andersonReversetimeDiffusionEquation1982}.

In this section, we present the theoretical formulation for generative modeling using discrete-state Markov processes. Detailed derivations can be found in the Appendices. We present the forward and backward Kolmogorov equations (Sec.~\ref{sec:forwardBackward}), and reverse-time processes (Sec.~\ref{sec:reverse}) for discrete-space diffusion modeling. We further elaborate (Sec.\ref{sec:asymptotic}) on the continuum limit of discrete-state systems, providing the connection to continuous-state systems, and describing the discrete-state score functions.  The loss functions (Sec.~\ref{sec:likelihood}) and sample generation (Sec.~\ref{sec:generation}) procedures are also provided. 

\subsection{Notations} \label{sec:notation}

\noindent{\bf State space.} We consider an $N$-dimensional state space $\Omega^N$, where each dimension of the data lives in a finite discrete-state space $\Omega$, and each state is labeled and ordered by $\left\{0\ldots M\right\}$. For example, in CIFAR-10, $N=32\times 32\times 3$ and $\Omega=\left\{0,1,\ldots M=255\right\}$. A state in this space, a high-dimensional random variable, is denoted by $\mathbf{X} \in \Omega^N$. For most of the derivations, it suffices to consider a single dimension (one color channel of a particular pixel) which will be denoted by $X\in \Omega$. We will use the lower-case symbols (e.g., $m$, $n$, $o$) to denote the dummy state variable. We will use the subscript to denote the time parameter of the random process: For discrete-time systems, the random process is denoted as $\mathbf{X}_k$, $k \in \mathbb{Z}_{\ge 0}$, and continuous-time random process is denoted as $\mathbf{X}_t$, $t\ge 0$.

\noindent{\bf Probability distributions.} We use $p$ to denote joint or conditional probabilities. In addition, a shorthand notation will be employed, for example with times $s,\,t\ge 0$, 
\begin{subequations}
  \begin{align}
    p_{(m,t)} :={}& \mathbb{P}\left\{X_t = m\right\},  \\
    p_{(n,t) \vert (m,s)} :={}& \mathbb{P}\left\{X_t = n \vert X_s = m\right\}, \\
    p_{(m,s),(n,t) \vert (o,0)} :={}& \mathbb{P}\left\{ X_s = m, X_t = n\vert  X_0 = o\right\}.
  \end{align}
\end{subequations}

\subsection{Forward and Backward equations}\label{sec:forwardBackward}

The forward process is described by the Chapman--Kolmogorov equations \cite{kolmogoroffUberAnalytischenMethoden1931,vanKampen,gardinerStochasticMethodsHandbook2009}. We will consider a one-dimensional Markov process applied independently to each of the dimensions of the joint state $\mathbf{X}$. In the continuous-time setting, the forward equation can be written as a Master Equation \cite{vanKampen,gardinerStochasticMethodsHandbook2009,weberMasterEquationsTheory2017}: 
\begin{equation}
    \frac{\text{d}}{\text{d} t} p_{(m,t) \vert (o,0)}= \sum_{m'} L^\dagger_{mm'} p_{(m',t) \vert (o,0)},  \label{eq:cforward}
\end{equation}
where $L$  is the generator of the continuous-time and discrete-state stochastic process and $L^{\dagger}$ is the adjoint operator of $L$.

We first explain our results for a subset of Markov processes where the states are only allowed to transition between neighboring states, so that the generator in Eq.~\eqref{eq:cforward} have a tri-diagonal structure. For this subset of processes, it is convenient to use the \emph{step operators} \cite{vanKampen} to reformulate the forward equation \eqref{eq:cforward}. 
The step operators $\mathcal{E}_{\pm}$ are defined by $\mathcal{E}_{\pm} f(m) \triangleq f(m\pm 1)$ for any test function $f$. Employing the Einstein summation convention for the $\pm$ signs $\sigma\in \left\{+,-\right\}$, the forward equation \eqref{eq:cforward} can be expressed by 
\begin{equation}
    \frac{\text{d}}{\text{d} t} p_{(m,t) \vert (o,0)}  = \left( \mathcal{E}_\sigma^\dagger -1 \right) \left [\nu_\sigma (m)\, p_{(m,t) \vert (o,0)}\right]  \label{eq:cforward2}
\end{equation}
where $\nu_\sigma (m)$ is the transition rate from state $m$ to $m + \sigma \cdot 1$. 

The backward equation for the continuous-time process \eqref{eq:cforward2} is the Kolmogorov backward equation \cite{kolmogoroffUberAnalytischenMethoden1931,vanKampen,gardinerStochasticMethodsHandbook2009} for the transition probabilities from state $m$ at an earlier time $s\ge 0$ to state at $n$ at a later time $t$:
\begin{equation}
    -\frac{\text{d}}{\text{d} s} p_{(n,t) \vert (m,s)} = \nu_\sigma (m) \left( \mathcal{E}_\sigma -1 \right)  p_{(n,t) \vert (m,s)}.  \label{eq:cbackward}
\end{equation}

\subsection{Reverse-time process}\label{sec:reverse}

Here we prescribe the reverse-time stochastic process, analogous to the prescription of \citet{andersonReversetimeDiffusionEquation1982} which is applied for Gaussian diffusion. Formally, the reverse-time process describes how conditional probability $p_{(m,s)\vert (n,T), (o,0)}$, $0\le s \le T$, evolves reversely in time, which is necessary for designing the likelihood or loss function and for generation, thereby laying the foundation for Diffusion Model learning and inference. Here, we present the results, and leave the detailed derivations in Appendix \ref{app:reverseTime}.

For the continuous-time Markov process \eqref{eq:cforward2}, the reverse-time evolutionary equation for any time $0 \le s \le t$ reads
\begin{align}
     -\frac{\text{d}}{\text{d} s }{}&   p_{(m,s)\vert(n,t),(o,0)} =  \label{eq:creverse}\\
     {}& \left(\mathcal{E}_\sigma - 1 \right) \left [ \nu_\sigma\left(m_\sigma'\right) \frac{p_{(m'_\sigma,s) \vert (o,0)}}{p_{(m,s) \vert (o,0)}} p_{(m,s)\vert(n,t),(o,0)} \right]. \nonumber
\end{align}

Note that the reverse-time process is an explicitly time-dependent Markov process, whose transition rates depend on the forward solution $p_{(m,s)\vert (n,0)}$. This is analogous to the fact that the drift and diffusion of the reverse-time Gaussian diffusion processes depend on the forward solution \cite{andersonReversetimeDiffusionEquation1982}. Note also that a state $m$ later in time can transit to a state $m'$ at an early time only if there exists a transition $m'\rightarrow m$ defined in the forward process (see Remark 1 in Appendix \ref{app:reverseTime}). These expressions, derived from formal operator algebra, can be understood intuitively as a conditional Bayes formula (see Appendix \ref{app:reverseTime}). 

While Eqs.~\eqref{eq:cforward2}-\eqref{eq:creverse} are specific to tri-diagonal transitions, all of these results generalize to arbitrary transition matrices, as shown in Appendix \ref{app:generalization}, by writing the transition matrix as a sum over banded processes and exploiting linearity. We also remark that the theory applies to arbitrary discrete-state spaces. The state space does not have to be ordered, although it is ordered in our application of image dataset below. Finally, as illustrated in Appendix \ref{app:reverseTime}, there exists a corresponding theory for generic discrete-time and discrete-time stochastic processes. 

\begin{figure*}[ht!]
    \centering
    \includegraphics[clip=true, trim={2.7cm 0cm 2.8cm 0cm}, width=1.0\textwidth]{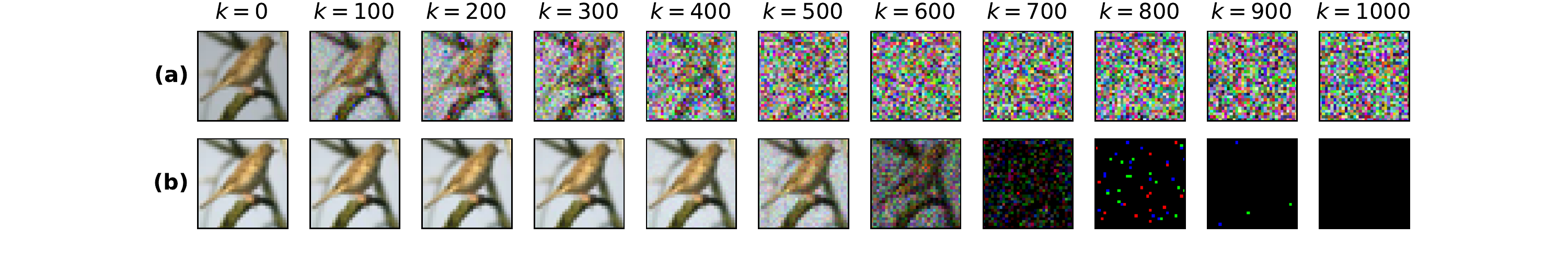}
    \caption{The forward process of (a) Gaussian diffusion and (b) Blackout Diffusion applied on an image, sampled at discrete time $k$. The colormap is adjusted per-image to better visualize the noisy signal of Blackout Diffusion at large times. Fig.~\ref{fig:imageDegradation_noScaling} shows images with unadjusted colormap.}
    \label{fig:imageDegradation}
\end{figure*}

\subsection{Relationship to Gaussian diffusion}\label{sec:asymptotic}

To shed light on the connection between our discrete-state diffusion formulas and Gaussian diffusion, we use Kramers--Moyal expansion \cite{vanKampen,gardinerStochasticMethodsHandbook2009,weberMasterEquationsTheory2017}, a standard asymptotic analysis to study the limiting behavior in a state space with large number discrete states ($M\gg1$) densely distributed in $\Omega$. Appendix \ref{app:asymptotics} presents the formal expansion, showing that both the forward and reversal discrete-state processes asymptotically converge to the standard (multiplicative) Gaussian diffusion in this limit. Importantly, the asymptotic expansion of the discrete-state reversal process is consistent with the reversal process (by \citet{andersonReversetimeDiffusionEquation1982}) of the asymptotically expanded forward process.

Performing the continuous limit also allows the identification of the discrete-state score function $s_\mathrm{dis,\sigma}$:
\begin{equation}
s_{\text{dis}, \sigma} \left(m, s\right)\propto \nu_\sigma\left(m'_\sigma\right)\frac{p_{(m'_\sigma,s)\vert (o,0)}-p_{(m,s)\vert(o,0)}}{p_{(m,s)\vert(o,0)}}, \label{eq:cts-score}
\end{equation}
which is an analog to the Stein score function, the key learning target for score-based generative models \cite{songGenerativeModelingEstimating2020,songImprovedTechniquesTraining2020,songScoreBasedGenerativeModeling2021a,songMaximumLikelihoodTraining2021}. Interestingly, there is a transformation to learn for each type of transition event into state $m$, rather than simply a transformation for each feature in the data. (see discussion in Appendices \ref{app:asymptotics} and \ref{app:generalAlgo}).

\subsection{Blackout Diffusion} \label{sec:pureDeathProcess}

Here, we choose a simple Markov process to illustrate the utility of the proposed discrete-state formulation. The process is adequate for the digital images, whose state space $\Omega = \left\{0\ldots 255\right\}$ is discrete and naturally ordered. One of the simplest processes of this class is the pure-death process, such that the only event is the transition $m \in \Omega $ to $m-1$ with a transition rate $\gamma m$:
\begin{equation}
    m \xrightarrow{\gamma m} m-1.  \label{eq:pureDeath}
\end{equation}
The pure-death process is a mathematical model for constant, independent decay, such as in radioactivity. For digital image data, it can be interpreted as such: Each unit intensity of a specific pixel and color channel decays from $1$ to $0$ with a rate $\gamma$. This process is a homogenous stochastic process, that is, the transition rate is constant in time. We can arbitrarily set the timescale such that $\gamma=1$. With Eq.~\eqref{eq:cforward}, the transition matrix can be defined by a banded $L^\dagger_{mm'} = m' \left(\delta_{m, m'-1} - \delta_{m, m'}\right)$, where $\delta$ is the Kronecker delta. Note that the pure-death process is distinct from Mask Diffusion \cite{austin21}, which has one-step transition from any state to the absorbing state, per pixel and per channel. For pure-death process, a state $n$ requires $n$ transition events before it reaches the absorbing state $0$, per pixel and per channel. The two processes are very similar on binary data, only differing in that Masked Diffusion is non-homogenous process (i.e., the transition rate is time-dependent). With the representation Eq.~\eqref{eq:cforward2}, we have the negative transition $\nu_-(m)= m$ and the positive transition $\nu_+(m)=0$, $m\in \Omega$.  It is elementary to derive the solution of the forward process as a binomial distribution\footnote{The pure-death process is a special case of the birth-and-death process or equivalently the M/M/1 queue, whose analytical solutions are known \cite{abateTransientBehaviorQueue1987,gardinerStochasticMethodsHandbook2009}.}, $X_t \sim \text{Binom}\left(X_0, e^{-t}\right)$, or equivalently
\begin{equation}
    p_{(m,t)\vert (o,0)} = \left(\begin{array}{c} o \\ m \end{array} \right) e^{- m t } \left(1-e^{-t}\right)^{\left(o-m\right) }. \label{eq:fowardSol}
\end{equation}

Thus, at any finite time $t\in \left(0, \infty\right)$, any initial state is diffused to a binomial distribution that decays exponentially. As $t\rightarrow \infty$, all initial states $o$ converge to the same state, $0$. Thus, the prior for this process consists of a $\delta$-distribution at a single point---an entirely black image---hence the name \emph{Blackout Diffusion}. Because the prior is singular\footnote{ The singular distribution is not a special property of discrete-state systems. There exists It\^o SDE that admits singular Diract $\delta$ distribution as their limiting distribution \cite{fellerTwoSingularDiffusion1951,fellerParabolic1952}.}, diffused samples cannot contain any information about the initial data, a point we will discuss further in Sec.~\ref{sec:discussion}. Examples of Gaussian and Blackout diffusion as forward processes are provided in Fig.~\ref{fig:imageDegradation}.
We will use the continuous-time formulation (i.e., Eqs.~\eqref{eq:cforward}, \eqref{eq:cbackward}, \eqref{eq:creverse}) because of the rather large state space $\left \vert \Omega \right \vert  = 256 \gg 1$. Preliminary experiments showed that using a discrete-time formulation would involve either too many time steps or too noisy transitions, resulting in lower-quality generated samples. 

\begin{figure*}[t!]
    \centering
    \includegraphics[clip=true, trim={2.7cm 0cm 2.8cm 0cm}, width=1.0\textwidth]{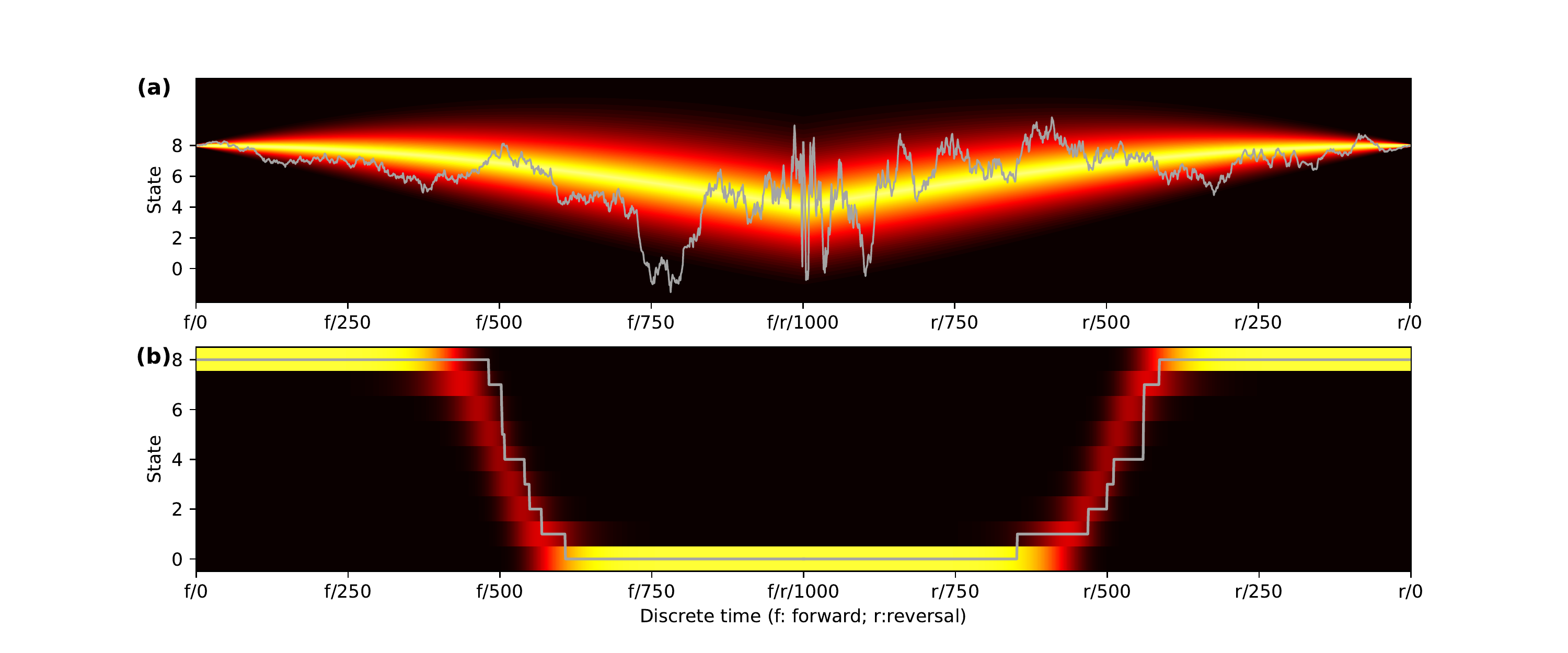}
    \caption{Probability distribution (heatmap) and paths (solid line) on a discrete-support $\Omega=\left\{0,1,\ldots M=8\right\}$. (a) Gaussian diffusion with de-quantization. The stochastic paths are not constrained on the discrete support, and also not bounded between $[0,M]$. Here, the Gaussian process is that in \cite{hoDenoisingDiffusionProbabilistic2020}, with the cosine schedule. The forward process brings all the states to a high-dimensional Gaussian. (b) The discrete-state pure-death process takes place only on $\Omega$ and bring all the states to a $\delta$-distribution at $0$. Here, we used the discrete observation times specified in Appendix \ref{app:schedule}. Both processes have $T=1000$ steps.}
    \label{fig:paths}
\end{figure*}

 Prescribed by Eq.~\eqref{eq:creverse}, the reverse-time dynamics is a birth-only process, with a non-trivial transition rate
\begin{equation}
    \nu_-\left(m'\right) \frac{p_{(m',s) \vert (o,0)}}{p_{(m,s) \vert (o,0)}} = \left(o-m\right) \frac{e^{-t}}{1-e^{-t}}. \label{eq:rRates}
\end{equation}
Figure~\ref{fig:paths} contrasts the forward and reverse-time processes for Gaussian Diffusion with the improved noise schedule \cite{nicholImprovedDenoisingDiffusion2021} and Blackout Diffusion.

Note that Eq.~\eqref{eq:rRates} prescribes only the instantaneous time-dependent transition rates. Due to the simplicity of Blackout Diffusion, we can establish that the solution of the reverse-time process after a finite-time leap, i.e., $p_{(m,s)\vert (n,t),(o,0)}$, $0 \le s \le t$, is also binomial in nature (see Appendix \ref{app:binomialBridge}):
\begin{align}
    p_{(m,s)\vert (n,t),(o,0)} ={}&  \left(\begin{array}{c} o-n \\ m-n \end{array} \right) r^{m-n} (1-r)^{o-m}, \nonumber \\
    r:={}& s\left(e^{-s}-e^{-t}\right)/\left(1-e^{-t}\right). \label{eq:binomialBridge}
\end{align}
In more compact notation, this can be written $X_s\vert (X_0, X_t) - X_t \sim  \text{Binom}\left(X_0-X_t, r\right)$. We label Eq.~\eqref{eq:binomialBridge} as the \emph{Binomial bridge formula}, since it is analogous to the key Brownian bridge formula used in Gaussian Generative Diffusion Models. Since we chose an independent decay process for each feature in the image data, all of the above can be immediately transferred from pixels $X \in \Omega$ to full images $\mathbf{X} \in \Omega^N$.

\subsection{Diffusion Schedule and Loss functions} \label{sec:likelihood}

For training, a set of $T$ \emph{observation times} $t_k \in \mathbb{R}_{\ge 0}$, $k=1,\ldots T$ is defined \emph{a priori}, ordered by $0 < t_1 < t_2 \ldots t_{T}$. In addition, $t_0:=0$. The choice of the observation times is analogous to the noise schedule in Gaussian diffusion models. Rather than treating the observation times/noise amplitudes as a hyperparameter to be optimized, as in \citet{sohl-dicksteinDeepUnsupervisedLearning2015,nicholImprovedDenoisingDiffusion2021,hoDenoisingDiffusionProbabilistic2020}, we take a mathematical approach to defining the schedule of observation times $t_k$. The schedule is designed to uniformize the change in the Fisher information of the degraded samples, and this is derived in Appendix \ref{app:schedule}. 

Similar to other diffusion models, the goal is to train a neural network which transforms samples at $t_{k}$ to $t_{k-1}$. However, we accomplish this with a novel approach: Observing the instantaneous transition rate Eq.~\eqref{eq:rRates} and the finite reverse-time propagation Eq.~\eqref{eq:binomialBridge}, this goal can be achieved by training the neural net to predict the difference $\mathbf{X}_0-\mathbf{X}_{t_{k-1}}$, given a realization of $\mathbf{X}_{t_{k}}$. We will write the neural-net predicted difference as $\mathbf{y}:=\mathbf{NN}\left(\mathbf{X}_{t_k}, t_k\right)$. We remark that this computational instantiation is drastically different from existing discrete diffusion implementations \cite{austin21,campbell2022} which model transition rates. With Blackout Diffusion, transition rates decays several orders of magnitudes, making it difficult to train a neural network to learn the transition rates. The remaining task is to establish the loss function that minimizes a metric on Eqs.~\eqref{eq:rRates} and \eqref{eq:binomialBridge}.

We use a likelihood approach to define the loss function. Appendix \ref{app:likelihood} presents the derivation, showing that for a sample $\mathbf{X}_{t_k}$ generated by a particular training sample $\mathbf{X}_0$, the loss function can be defined as:
\begin{equation}
l_i = \left(t_k - t_{k-1}\right) e^{-t_{k}}\left[ y_i - \left(\mathbf{X}_0-\mathbf{X}_{t_{k}}\right)_i \log y_i \right],\label{eq:lossInstantaneous}
\end{equation}
where $i$ indexes components of the multidimensional state. A finite-time correction by the binomial bridge formula is 
\begin{align}
l_i ={}& \left(e^{-t_{k-1}}-e^{-t_k}\right) \left[ y_i  - \left(\mathbf{X}_0-\mathbf{X}_{t_k}\right)_i \log y_i \right]. \label{eq:lossFinite}
\end{align}
We average over each component $i$, samples in the training set $\mathbf{X}_0$, and randomly sampled observation time index $k$; the full loss is $\mathcal{\ell} = \left \langle l_i\right\rangle_{\{\mathbf{X_0},k,i\}}$. 
Note that if one chooses a uniformly distributed observation times and if the time difference $\Delta := t_k-t_{k-1} \ll 1$, Eq.~\eqref{eq:lossFinite} converges asymptotically to Eq.~\eqref{eq:lossInstantaneous}.

\subsection{Generating sample images} \label{sec:generation}

For image generation, we chose $t_T=15$, a time that is deemed long enough to approximate $t\rightarrow \infty$, in which limit all initial states converged to 0 (i.e., all the images degrade to all black). More accurately speaking, $p_{(0,15)\vert (255,0)} > 0.9999$. To generate samples, we start with an all-black image, use the trained neural network to predict $\mathbf{X}_{t_0} - \mathbf{X}_{t_{T}}$, clipped at 0 and 255. Then, we can either use the general $\tau$-leaping algorithm \cite{tauLeaping21} (which is analogous to the Euler--Maruyama integrator for It\^o SDEs \cite{kloedenNumericalSolutionStochastic1999}) to generate a Poisson random number, or exploit the analytically exact binomial bridge Eq.~\eqref{eq:binomialBridge}, whose applicability is limited to Blackout Diffusion. The former sampling approach is similar to integrating It\^o SDE in \cite{songScoreBasedGenerativeModeling2021a}, and the latter approach is analogous to the Denoising Diffusion Probabilistic Model \cite{hoDenoisingDiffusionProbabilistic2020}. The drawn numbers are added to the all-black image to obtain $\mathbf{X}_{t_{T-1}}$. The procedure repeats until $\mathbf{X}_{0}$ is obtained.  

We summarize the proposed training and inference methods in Algorithms \ref{alg:training} and \ref{alg:inference}. The algorithms can be generalized to accommodate arbitrary continuous-time and discrete-state Markov process in Appendix \ref{app:generalAlgo}.

\begin{algorithm}[tb]
   \caption{Training Blackout Diffusion}
   \label{alg:training}
\begin{algorithmic}
   \REPEAT
   \STATE $\mathbf{X}_0 \leftarrow \mathbf{x}$, drawn from the training set
   \STATE Draw an index $k$ from $\left\{1, \ldots T\right\}$ uniformly
   \STATE $\mathbf{X}_t \sim \text{Binomial}\left(\mathbf{X}_0, e^{-t_k}\right)$ (element-wise)
   \STATE $\mathbf{y} \leftarrow \mathbf{NN}_\theta \left(\mathbf{X}_{t_k}, k\right)$ 
   \IF{Using instantaneous loss function Eq.~\eqref{eq:lossInstantaneous}}
   \STATE $\omega_k \leftarrow \left(t_k -t_{k-1}\right) e^{-t_{k}}$ 
   \ELSIF{Using finite-time loss function Eq.~\eqref{eq:lossFinite}}
   \STATE $\omega_k \leftarrow e^{-t_{k-1}} - e^{-t_{k}}$
   \ENDIF
   \STATE $l \leftarrow \omega_k \times \text{mean} \left\{ \mathbf{y} - \left(\mathbf{X}_0-\mathbf{X}_{t_k}\right)\log \mathbf{y}\right\}$
   \STATE Take a gradient step on $\nabla_\theta l$
   \UNTIL{Converged}
\end{algorithmic}
\end{algorithm}

\begin{algorithm}[tb]
   \caption{Generating images by Blackout Diffusion}
   \label{alg:inference}
\begin{algorithmic}
   \STATE Initiate a blacked-out image $\mathbf{X}_{t_{T}}=0$ 
   \FOR{$k=T$ {\bfseries to} $1$}
   \STATE $\mathbf{y}_\theta \leftarrow \text{clip} \left( \mathbf{NN}_\theta \left(\mathbf{X}_{t_k}, k\right), \,\mathbf{0},\, 255\mathbf{I}-\mathbf{X}_{t_k}\right) $ 
   \IF{Using Poisson}
   \STATE $\mathbf{d} \sim \text{Poisson}\left( \mathbf{y}_\theta \frac{e^{-t_k}}{1-e^{-t_k}}\right)$ (element-wise)
   \STATE $\mathbf{X}_{t_{k-1}} \leftarrow \text{clip}\left(\mathbf{X}_{t_{k}}+\mathbf{d}, \mathbf{0}, 255 \mathbf{I} \right)$
   \ELSIF{Using Binomial Bridge}
   \STATE $\mathbf{d} \sim \text{Binomial}\left(\mathbf{y}_\theta, \frac{e^{-t_{k-1}}-e^{-t_k}}{1-e^{-t_k}}\right)$ (element-wise)
   \STATE $\mathbf{X}_{t_{k-1}} \leftarrow \mathbf{X}_{t_{k}}+\mathbf{d}$
   \ENDIF
   \ENDFOR
\end{algorithmic}
\end{algorithm}

\begin{table}[t]
\caption{Results of the numerical experiments (iteration 300K).}
\label{table:unique}
\vskip 0.15in
\begin{center}
\begin{small}
\begin{sc}
\begin{tabular}{cccc}
\toprule
    Loss  &  Sampling &  FID & IS \\
\midrule
    Instantaneous &  Binomial  &   4.77 & 9.01 \\
    Instantaneous  &  Poisson  &  4.92 &  9.18 \\
    Finite-time &  Binomial  & 4.83 & 9.00 \\
    Finite-time &  Poisson  &  5.01 & 9.08\\
\bottomrule
\end{tabular}
\end{sc}
\end{small}
\end{center}
\vskip -0.1in
\end{table}

\begin{table*}[]
\caption{Summary of discrete methods} \label{table:not-so-unique}

\begin{center}
\begin{small}
\begin{sc}
\begin{tabular}{@{}cccccc@{}}
\toprule
Dataset & Discrete Methods             & \begin{tabular}[c]{@{}c@{}}Starts from a \\ singular point?\end{tabular} & \begin{tabular}[c]{@{}c@{}}Corrector \\ steps\end{tabular} & \begin{tabular}[c]{@{}c@{}}Time \\ homogeneous\end{tabular}   & FID (↓) \\ \midrule[0.9pt]
        & D3PM Absorbing  \cite{austin21}               & $\checkmark$                  & None                   & $\times$         & 30.97   \\
        & D3PM Gauss \cite{austin21}                   & $\times$                      & None                   & $\times$         & 7.34    \\
{\scriptsize CIFAR10} & $\tau$LDR-0 \cite{campbell2022}                  & $\times$                      & None                   & $\times$         & 8.10    \\
        & $\tau$LDR-10 \cite{campbell2022}                 & $\times$                      & 10                     & $\times$         & \textbf{3.74}    \\
        & BlackOut (Ours) & $\checkmark$                  & None                   & $\checkmark$     & 4.77    \\
        &            BlackOut (Ours)                      & $\checkmark$                  & 2                      & $\checkmark$     & 4.58    \\ \midrule
{\scriptsize Binarized MNIST}  & BlackOut (Ours)                  & $\checkmark$                  & None                   & $\checkmark$     & \textbf{0.02}    \\ \midrule
{\scriptsize CelebA}  & BlackOut (Ours)                  & $\checkmark$                  & None                   & $\checkmark$     & \textbf{3.21}   \\ \bottomrule
\end{tabular}

\end{sc}
\end{small}
\end{center}

\end{table*}

\section{Numerical Experiments} \label{sec:numerics}

We chose the CIFAR-10 dataset to validate the feasibility of using Blackout Diffusion for generative modeling. We stress that our goal is to provide evidence that the theoretical results for discrete-state diffusion, derived above, are useful for practical tasks, rather than to advance the state-of-the-art\footnote{By the time the authors concluded this write-up the lowest FID (1.97) was attributed to \cite{cifar_SOTA}} for CIFAR-10 or any other dataset. To do so, we adopted the improved Noise Conditional Score Network (NCSN++) architecture presented in \citet{songScoreBasedGenerativeModeling2021a,scorenet} to serve the $\mathbf{NN}$ function in Eqs.~\eqref{eq:lossInstantaneous} and \eqref{eq:lossFinite}. We made two minimal modifications on NCSN++ which bring the architecture in line with the theoretical requirements of Blackout Diffusion Modeling. (1) We applied a softplus function on the output of NCSN++ to ensure the positivity of the output. This is because the target ($\mathbf{X}_0-\mathbf{X}_{t_k} \ge 0$) is non-negative in Blackout Diffusion. (2) For inference, we clip the output of $\text{softplus}\left(\text{NCSN++}\left(\cdot \right) \right)$ by 0 and $255 \mathbf{I}-\mathbf{X}_{t_k}$, followed by a round-off to the nearest integer. We used mini-batches with $256$ samples, and the training was stopped at {300K} iterations, above which we observed degraded quality of the generated samples. We did not modify other hyperparameters of the NSCN++. We fixed $T=1000$ in this feasibility experiment. We performed the analysis on both loss functions, Eqs.~\eqref{eq:lossInstantaneous} and Eq.~\eqref{eq:lossFinite}, and by both binomial and Poisson sampling. The training was carried out by two NVIDIA A100 GPUs for $\sim$72 hr. Due to limited computational resources, we only trained once for each loss function. We then generated 50K samples, which took approximately 10.5 hours. 

\begin{figure*}[th!]
    \centering
    \includegraphics[clip=true, trim={2.7cm 0cm 2.8cm 0cm}, width=1.0\textwidth]{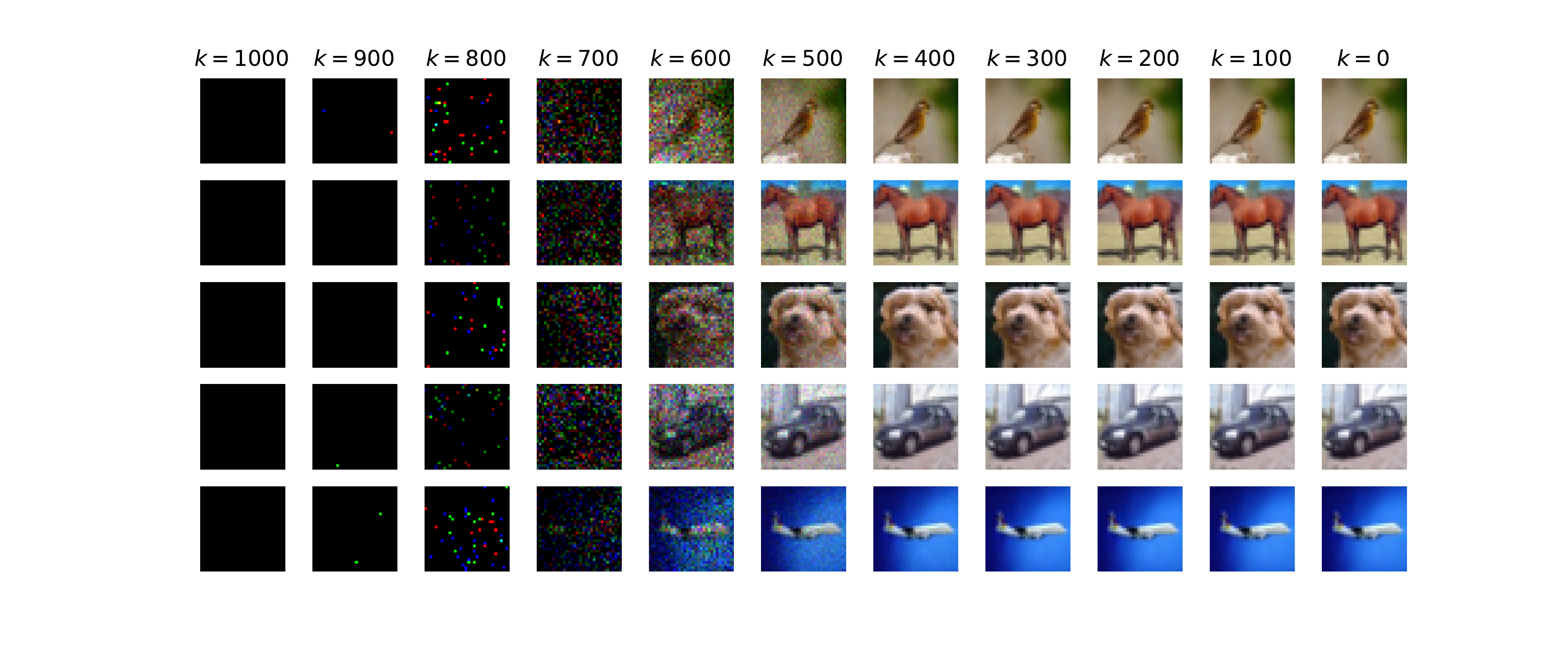}
    \caption{Image generation based on Blackout Diffusion. The color is adjusted per-image to better visualize the noisy signal at large times. Fig.~\ref{fig:imageGeneration_noScaling} shows images with unadjusted colormap.}
    \label{fig:imageGeneration}
\end{figure*}

Selected samples are visualized in Fig.~\ref{fig:imageGeneration}, which shows how the Blackout Diffusion Model generates different images from the same terminal condition ($\mathbf{X}_{t_T}=0$). Using 50K generated images, we computed the Fr\'echet Inception Distance (FID) to the training set and the Inception score (IS) in Table \ref{table:unique}. Our results suggest that the quality of the generated samples is not sensitive to the choice of the loss function, and using the binomial bridge formula for sampling is slightly advantageous. We visualize 400 generated images for each case in Figs.~\ref{fig:grid-f-b}-\ref{fig:grid-i-p}. Results of binary MNIST and CelebA are shown in Table \ref{table:not-so-unique} and Figs.~\ref{fig:mnist-generation}-\ref{fig:celebA}.

\section{Discussion and Conclusion} \label{sec:discussion}

We developed an exact theoretical framework for general discrete-state Markov processes, on either discrete- or continuous-time domain, which enables the exploration of Diffusion Modeling using a large class of stochastic processes. Our theoretical formulation for the discrete-state processes is analogous to \citet{andersonReversetimeDiffusionEquation1982} for Gaussian processes\footnote{We emphasize that one should not confuse the reverse-time process, our Eq.~\eqref{eq:creverse}, with the Kolmogorov backward equation \eqref{eq:cbackward}. While \citet{sohl-dicksteinDeepUnsupervisedLearning2015} attributed the credit to \citet{fellerTheoryStochasticProcesses1949}, which presents the standard Kolmogorov backward equations \cite{kolmogoroffUberAnalytischenMethoden1931}, \citet{andersonReversetimeDiffusionEquation1982} derived the general \textit{reversal} stochastic process which enables the generative diffusion modeling by Gaussian diffusion.}. To the authors' knowledge, a complete analogous theory for general discrete-state Markov processes has not yet been proposed.  These theoretical results can be summarized as such: (1) We generalized the diffusion models based on those SDEs whose underlying noise is Wiener process \cite{songScoreBasedGenerativeModeling2021a} to discrete-state Markov processes. Besides yielding Blackout Diffusion based on the pure-death process, this paves the way to include more complex stochastic processes that involve some discreteness, e.g., jump-diffusion model \cite{Huang_2014}, piecewise-deterministic Markov process \cite{faggionatoNonequilibriumThermodynamicsPiecewise2009}, Gaussian diffusion with Markov switching \cite{mao2006}. Some of these more complex models could expand the current extent of generative diffusion models. (2) We derived the discrete-state score functions, Eqs.~\eqref{eq:cts-score} and \eqref{eq:dis-score}. These functions are analogous to the Stein score function \cite{songGenerativeModelingEstimating2020}. Since the score function was not used for training Blackout Diffusion, it remains for future work to see if score-based Diffusion Modeling \cite{songScoreBasedGenerativeModeling2021a,songGenerativeModelingEstimating2020} can be applied to discrete-state processes. For generic discrete-state models, the reverse-time generator Eq.~\eqref{eq:creverse} holds the key to enable FHDM (our Eq.~\eqref{eq:creverse} is the discrete-state counterpart of the continuous-state Itô SDE (9) in \citet{Ye2022}). FHDM uses distinct training and sampling methods based on Doob's transformation to achieve improved quality and efficiency. Although in this paper we did not combine our theoretical formulation and FHDM, such a combination can be a fruitful research direction. Furthermore, as established in \citet{Ye2022}, FHDM can be connected to the (continuous-state) Schr\"odinger--F\":ollmer bridge formalism \citep{vargas2021a,vargas2021b,deBortoli2021} and the (continuous-state) Path Integral Sampler \citep{PIS2022}, we hypothesize that a successful application of our theoretical formulation to FHDM could initiate a new class of problems on corresponding discrete-state Schr\"odinger--F\"ollmer bridge and Path Integral formalism, which to our best knowledge do not exist yet. (3) We associate the ``noise schedules'' to the observation times of a continuous-time diffusion process. Such an association allowed us to design a principled, albeit heuristic, argument to construct the observation times based on Fisher information of the forward process. We remark that a different heuristic existed for training Mask Diffusion \cite{austin21}. (4) Using the exact theoretical formulation, we formulated a loss function based on maximum likelihood. This contrasts to \citet{campbell2022}, who took an approximate (variational) approach and then send the discrete time step down to zero. Our result (in Appendix \ref{app:generalAlgo}) shows that the variational bound in \citet{campbell2022} is tight in the continuum-time limit. Finally, (4) we provide a demonstration that discrete-state Diffusion Modeling can be effective while exhibiting completely different qualitative behavior from Gaussian diffusion, by learning a Blackout Diffusion model on binarized MNIST, CIFAR-10, and CelebA-64.

These results also allow us to address conceptual questions about the technique of Diffusion Modeling more broadly:

\emph{Is Gaussian noise special?} Our numerical experiment on Blackout Diffusion shows that Gaussian noise is not a crucial property for generative diffusion modeling. The discrete-state (Poissonian) noise in Blackout Diffusion can also be used for Generative Diffusion Modeling. This discovery is in line with a few other recently proposed models, e.g., Cold Diffusion \citet{Bansal2022ColdNoise} and Multinomial Diffusion \citet{hoogeboom2021argmax,song2020denoising}. Beyond the computational experiments, the existence of the generic instantaneous loss formula implies that the learning is tractable for any discrete-state process. \\*[4pt]
\emph{Must the forward and reverse processes correspond to noisifying and de-noising, respectively?} Often, the forward process is described as one adds noise to the image \cite{hoDenoisingDiffusionProbabilistic2020,hoogeboom2021argmax}. Blackout Diffusion highlights that this association must be taken with some subtlety. If noise corresponds to variance in the data across the space of samples of the stochastic process, for $t<\log 2$, the forward process for Blackout Diffusion adds noise, but for $t>\log 2$, the forward process removes noise as samples converge towards a blank image. The scenario is, of course, the reverse in the process of generating an image. As such, the first phase of inference involves adding noise to the sample. However, the whole forward (respectively, reverse) Blackout Diffusion is a degradation (reconstruction) process that removes (adds) information of the generated samples. This represents an apparent paradox: information is constantly diffusing away, despite the fact that all the pixels are converging to the same singular distribution, which has maximal information in the sense of Shannon entropy. As such, to declare whether a process is adding noise or removing noise critically depends on what type of stochastic process is used, and how noise is defined. Notably, the same phenomenon can be observed in Mask Diffusion, although the ``denoisng'' nature was not discussed \cite{austin21}. In light of this, we find it more accurate to simply declare that the forward and reverse processes are stochastic processes.\\*[4pt]
\emph{Does the final state $\mathbf{X}_{t_T}$ constitute a latent-space representation?} The final state of the forward process is sometimes conceived as a latent-space representation, e.g., see \citet{songScoreBasedGenerativeModeling2021a}. Under this view, Diffusion Modeling appears conceptually very similar to Variational Autoencoders \cite{kingmaAutoencodingVariationalBayes2014} or Normalizing Flows \cite{rezende2015variational}. However, both Mask Diffusion \cite{austin21} and Blackout Diffusion converges all data to a single discrete point. Therefore, it seems impossible to conceive of the terminal space of the forward process as constituting a latent representation of the data. This highlights a fundamental difference between Diffusion Modeling and deterministic generative modeling. We argue that $\mathbf{X}_{t_T}$ is not a latent space, but the path-based conception of \citet{rombach2022high}, where  ($\mathbf{X}_{t_{1\ldots T-1}}$) may be viewed as a latent representation of the data---although this joint space is far larger than the original state space. With the similar spirit, we argue that the Cold Diffusion \cite{Bansal2022ColdNoise} is more a flow model than a diffusion model because it cannot be directly applied to singular priors. 

Finally, we would like to remark on the motivations that gave rise to our results and on future work that might be built from them. These results, collectively, arose out of the Authors' endeavor to understand the theory behind Diffusion Modeling---like many Machine Learning advancements, a fairly complex foundational idea yields, in practice, a few simple formulas to implement. A great many recent works have treated the exploration of Diffusion Modeling as a problem of hyperparameters. Practical but \emph{ad hoc} processes such as dequantization/requantization, and noise schedules have been included in pursuit of an engineering goal to improve model performance, but these stray significantly from the theoretical framework governing the underlying learning task. However, when looking at fundamentally different data domains---including but not limited to the discrete-space examples laid out in our introduction---this line of attack will likely introduce even machine learning algorithms for which the success cannot be explained and replicated. We believe that \emph{ad hoc} procedures which improve performance are less likely to transfer across tasks or even individual datasets. In contrast, when a clear theoretical framework is provided, the individual pieces of the learning task, such as the forward evolution, reverse process, and loss function, can be more closely examined and tested, and the exploration of innovations can be driven by understanding in addition to empirical testing.

Having laid out the exact theoretical framework, discrete-space Diffusion Modeling thus presents many well-motivated opportunities. For example, we hypothesize that with further work, discrete-space Diffusion Models may be found that are far more computationally efficient than Gaussian Diffusion Models, because the state space is far smaller (mathematically, infinitely so) than in the continuous case. An empirical hint that this may be so comes from the fact that we adopted an existing continuous-space architecture (NCSN++) out-of-the-box and achieved reasonable results; further search of the space of network architectures and hyperparameters ought to yield some improvement. 

Another advantage of our framework is that the instantaneous loss, Eq.~\eqref{eq:lossInstantaneous}, can be applied without any bridge formula that prescribes the reverse process analytically. Yet another advantage is in the computational tractability of the forward solution. For Gaussian Diffusion, the forward process is only computationally tractable because it can be analytically solved, and other continuous-state process which cannot be analytically solved are not computationally tractable. However, the forward evolution of discrete-state is very often numerically tractable, ask it involves powers (discrete-time) or exponentiation (continuous-time) of the transition matrix. These can be efficiently computed, e.g.~by the Krylov subspace method \cite{saad92,gaudreault2018kiops}, bypassing the need for an analytical forward solution. Because of both of these advantages, we anticipate a rich library of discrete-state processes that can be used for diffusion modeling, and hope that these processes can be tailor-matched to the rich space of discrete-state data and systems that arise in the natural world. 

\section*{Reproducibility and code availability}
The codes we developed to perform the experiments are deposited at \url{https://github.com/lanl/Blackout-Diffusion}, with a C-number C23047 approved by the Richard P.~Feynman Center for Innovation (FCI) at the Los Alamos National Laboratory.

\section*{Acknowledgements} \label{sec:acknowledgements}
YTL is supported by the Laboratory Directed Research and Development (LDRD) Project ``Uncertainty Quantification for Robust Machine Learning'' (20210043DR). ZRF and JES are supported by the Center for the Nonlinear Studies through the LDRD. NL acknowledges the support of LDRD project 20230290ER. The authors acknowledge significant support from the Darwin test bed at Los Alamos National Laboratory (LANL), funded by the Computational Systems and Software Environments subprogram of LANL's Advanced Simulation and Computing program. ZF notes that this manuscript has been authored by UT-Battelle, LLC under Contract No.~DE-AC05-00OR22725 with the U.S. Department of Energy. YTL dedicates this paper to the memory of his mentor Prof.~Charles R. Doering, whose invaluable knowledge and techniques greatly influenced this work. Despite Charlie left this world prematurely, his teachings will continue to inspire future research.

\newpage
\bibliography{refs}
\bibliographystyle{icml2023}


\appendix
\onecolumn
\section{Deriving the Reverse-Time Evolutions} \label{app:reverseTime}

We first derive the reverse-time evolution for the continuous-time process \eqref{eq:cforward2}. With $0\le s \le t$, recall that the conditional probabilities
\begin{equation}
    p_{(m,s),(n,t)\vert (o,0)} = p_{(n,t) \vert (m,s), (o,0)} p_{ (m,s)\vert (o,0)} = p_{(n,t) \vert (m,s)} p_{ (m,s)\vert (o,0)}.
\end{equation}
We use the Markov property $p_{(n,t) \vert (m,s), (o,0)}=p_{(n,t) \vert (m,s)}$ to establish the last equality. Applying $-\text{d}/\text{d}s $ to the equation above and by the chain rule, we arrive at
\begin{align}
    -\frac{\text{d}}{\text{d} s} p_{(m,s),(n,t)\vert (o,0)}  ={}&  -\frac{\text{d}  p_{(n,t) \vert (m,s)}}{\text{d} s} p_{ (m,s)\vert (o,0)} -p_{(n,t) \vert (m,s)} 
 \frac{\text{d} p_{ (m,s)\vert (o,0)} }{\text{d} s} \nonumber \\
 ={}& \nu_\sigma (m) p_{ (m,s)\vert (o,0)} \left( \mathcal{E}_\sigma -1 \right)  p_{(n,t) \vert (m,s)}  - p_{(n,t) \vert (m,s)} \left( \mathcal{E}_\sigma^\dagger -1 \right) \left [\nu_\sigma (m)\, p_{(m,s) \vert (o,0)}\right]\label{eq:alice}
\end{align}
by the forward and backward equations \eqref{eq:cforward2} and \eqref{eq:cbackward}. Next, we apply the product identity\footnote{Consider two test functions $f_1$ and $f_2$ of $m$. $\left(\mathcal{E}_\sigma^\dagger-1\right) \left[f_1\left(m\right) f_2\left(m\right)\right] = f_1\left(\mathcal{E}_\sigma^\dagger\left[m\right]\right) f_2\left(\mathcal{E}_\sigma^\dagger\left[m\right]\right) - f_1(m)f_2(m) $. The product rule can be established by adding and subtracting a term $f_1\left(\mathcal{E}_\sigma^\dagger\left[n\right]\right) f_2\left(m\right)$, which leads to $\left(\mathcal{E}_\sigma^\dagger-1\right) \left[f_1\left(m\right) f_2\left(m\right)\right] = f_1\left(\mathcal{E}_\sigma^\dagger \left[m\right]\right) \left[\left(\mathcal{E}_\sigma^\dagger-1\right) f_2 \left(m\right)\right] + f_2\left(m\right) \left[\left(\mathcal{E}_\sigma^\dagger-1\right) f_1 \left(m\right)\right]$.} to the terms involving the forward equation
\begin{align}
    - p_{(n,t) \vert (m,s)} \left( \mathcal{E}_\sigma^\dagger -1 \right) \left [\nu_\sigma (m)\, p_{(m,s) \vert (o,0)}\right] ={}& -\left(\mathcal{E}_\sigma^\dagger -1 \right) \left[p_{(n,t) \vert (m,s)} \nu_\sigma (m)\, p_{(m,s) \vert (o,0)} \right] \nonumber \\{}& + \left\{\mathcal{E}_\sigma^\dagger\left[ \nu_\sigma\left(m\right) p_{(m,s) \vert (o,0)} \right]\right\} \left\{\left(\mathcal{E}_\sigma^\dagger -1 \right) p_{(n,t) \vert (m,s)} \right\},
\end{align}
to expand Eq.~\eqref{eq:alice}:
\begin{align}
    -\frac{\text{d}}{\text{d} s} p_{(m,s),(n,t)\vert (o,0)}  ={}& 
    \nu_\sigma (m) p_{ (m,s)\vert (o,0)}  p_{(n,t) \vert (\mathcal{E}_\sigma\left[m\right],s)}
    - \nu_\sigma (m) p_{(n,t) \vert (m,s)} p_{ (m,s)\vert (o,0)} \nonumber \\
    {}& -\left(\mathcal{E}_\sigma^\dagger -1 \right) \left[p_{(n,t) \vert (m,s)} \nu_\sigma (m)\, p_{(m,s) \vert (o,0)} \right] \nonumber \\
    {}& + \nu_\sigma\left(\mathcal{E}_\sigma^\dagger\left[m\right]\right) p_{(\mathcal{E}_\sigma^\dagger\left[m\right],s) \vert (o,0)} p_{(n,t) \vert (\mathcal{E}_\sigma^\dagger\left[m\right],s)}
    \nonumber \\{}& - \nu_\sigma\left(\mathcal{E}_\sigma^\dagger\left[m\right]\right) p_{(\mathcal{E}_\sigma^\dagger\left[m\right],s) \vert (o,0)} p_{(n,t) \vert (m,s)}. \label{eq:bob}
\end{align}
Next, note that three of the terms in the RHS of Eq.~\eqref{eq:bob} cancel due to the identity
\begin{align}
\left(\mathcal{E}_\sigma^\dagger-1\right) {}& \left[\nu_\sigma (m) p_{(n,t) \vert (m,s)} p_{ (m,s)\vert (o,0)} \right] = \nonumber \\
{}&\nu_\sigma\left(\mathcal{E}_\sigma^\dagger\left[m\right]\right) p_{(\mathcal{E}_\sigma^\dagger\left[m\right],s) \vert (o,0)} p_{(n,t) \vert (\mathcal{E}_\sigma^\dagger\left[m\right],s)} - \nu_\sigma (m) p_{(n,t) \vert (m,s)} p_{ (m,s)\vert (o,0)},
\end{align}
Eq.~\eqref{eq:bob} can be simplified 
\begin{align}
    -\frac{\text{d}}{\text{d} s} p_{(m,s),(n,t)\vert (o,0)} ={}& 
    \nu_\sigma (m) p_{ (m,s)\vert (o,0)} p_{(n,t) \vert (\mathcal{E}_\sigma\left[m\right],s)} 
     - \nu_\sigma\left(\mathcal{E}_\sigma^\dagger\left[m\right]\right) p_{(\mathcal{E}_\sigma^\dagger\left[m\right],s) \vert (o,0)} p_{(n,t) \vert (m,s)}\nonumber \\
     ={}& \left(\mathcal{E}_\sigma - 1 \right) \left[\nu_\sigma\left(\mathcal{E}_\sigma^\dagger\left[m\right]\right) p_{(\mathcal{E}_\sigma^\dagger\left[m\right],s) \vert (o,0)} p_{(n,t) \vert (m,s)} \right],
\end{align}
by the identity of any step operators 
$\mathcal{E}_\sigma \mathcal{\mathcal{E}_\sigma^\dagger} =1$. Next, multiplying and dividing $p_{(m,s)\vert(o,0)} $ leads to the evolutionary equation of the joint ($(m,s)$ and $(n,t)$) conditional (on initial condition $(o,0)$) 
\begin{align}
     -\frac{\text{d}}{\text{d} s} p_{(m,s),(n,t)\vert (o,0)}
    ={}& \left(\mathcal{E}_\sigma - 1 \right) \left[\nu_\sigma\left(\mathcal{E}_\sigma^\dagger\left[m\right]\right) \frac{p_{(\mathcal{E}_\sigma^\dagger\left[m\right],s)\vert (o,0)}}{p_{(m,s)\vert(o,0)}} p_{(n,t) \vert (m,s)} p_{(m,s)\vert(o,0)} \right]\nonumber \\
    = {}& \left(\mathcal{E}_\sigma - 1 \right) \left[\nu_\sigma\left(\mathcal{E}_\sigma^\dagger\left[m\right]\right) \frac{p_{(\mathcal{E}_\sigma^\dagger\left[m\right],s)\vert (o,0)}}{p_{(m,s)\vert(o,0)}} p_{(n,t),(m,s)\vert (o,0)} \right]. \label{eq:charlie}
\end{align}
which is a closed-form evolution of the joint distribution $p_{(n,t), (m,s)\vert(o,0)}$, given the forward solution $p_{(m,s)\vert(o,0)}$. 
Finally, we divide both side of the equation on a particular terminal probability $p_{(n,t)\vert (o,0)}$ (assumed finite) to obtain Eq.~\eqref{eq:creverse}:
\begin{align}
     -\frac{\text{d}}{\text{d} s} p_{(m,s) \vert(n,t),(o,0)}
    =\left(\mathcal{E}_\sigma - 1 \right) \left[\nu_\sigma\left(\mathcal{E}_\sigma^\dagger\left[m\right]\right) \frac{p_{(\mathcal{E}_\sigma^\dagger\left[m\right],s)\vert (o,0)}}{p_{(m,s)\vert(o,0)}} p_{(m,s)\vert(n,t), (o,0)} \right]. \label{eq:david}
\end{align}

The derivation of the reverse-time evolution for the discrete-time process mostly follows the above derivation, but with a few additional technical details. With discrete-time index $k, \ell\in \mathbb{Z}_{\ge0}$, $k<\ell$, the forward and backward equations read
\begin{align}
   p_{(m,k+1) \vert (o,0)} ={}&  p_{(m,k) \vert (o,0)} + \left( \mathcal{E}_\sigma^\dagger -1 \right) \left [\mu_\sigma (m)\, p_{(m,k) \vert (o,0)}\right], \label{eq:dforward2} \\
   p_{(n,T) \vert (m,k)} ={}& p_{(n,T) \vert (m,k+1 )} + \mu_\sigma (m) \left( \mathcal{E}_\sigma -1 \right) p_{(n,T) \vert (m,k+1 )}, \label{eq:dbackward}
\end{align}
where $\mu_\sigma (m)$ is the transition probability from state $m$ to $m + \sigma \cdot 1$.
We first express the conditional probabilities $p_{(m,k),(n,\ell)\vert (o,0)}=p_{(n,\ell)\vert (m,k)} p_{(m, k)\vert (o,0)}$; again, the Markov property $p_{(n,\ell)\vert (m,k), (o, 0)}= p_{(n,\ell)\vert (m,k)}$ is used. Next, we establish an expression that is analogous to Eq.~\eqref{eq:alice}:
\begin{align}
    {}& p_{(m,k),(n,\ell)\vert (o,0)} - p_{(m,k+1),(n,\ell)\vert (o,0)} =  p_{(n,\ell)\vert (m,k)} p_{(m, k)\vert (o,0)} - p_{(n,\ell)\vert (m,k+1)} p_{(m, k+1)\vert (o,0)} \nonumber \\
    ={}& -  p_{(m, k)\vert (o,0)} \left[p_{(n,\ell)\vert (m,k+1)} - p_{(n,\ell)\vert (m,k)} \right] + p_{(n,\ell)\vert (m,k+1)} \left[ p_{(m, k)\vert (o,0)} - p_{(m, k+1)\vert (o,0)} \right] \nonumber \\
    ={}& p_{(m, k)\vert (o,0)} \mu_\sigma (m) \left( \mathcal{E}_\sigma -1 \right) p_{(n,\ell) \vert (m,k+1 )} - p_{(n,\ell)\vert (m,k+1)} \left( \mathcal{E}_\sigma^\dagger -1 \right) \left [\mu_\sigma (m)\, p_{(m,k) \vert (o,0)}\right], \label{eq:juliet}
\end{align}
where the last equality was established by the forward and backward equations, \eqref{eq:dforward2} and \eqref{eq:dbackward}. Next, we use the product identity to the terms involving the forward equation 
\begin{align}
    {}& - p_{(n,\ell)\vert (m,k+1)} \left( \mathcal{E}_\sigma^\dagger -1 \right) \left [\mu_\sigma (m)\, p_{(m,k) \vert (o,0)}\right] \nonumber \\= {}& - \left( \mathcal{E}_\sigma^\dagger -1 \right) \left [p_{(n,\ell)\vert (m,k+1)} \mu_\sigma (m)\, p_{(m,k) \vert (o,0)}\right] + \left[\mathcal{E}_\sigma^\dagger\left(\mu_\sigma (m)\, p_{(m,k) \vert (o,0)}\right)\right] \left[\left( \mathcal{E}_\sigma^\dagger -1 \right) p_{(n,\ell)\vert (m,k+1)} \right]
\end{align}
to expand \eqref{eq:juliet}:
\begin{align}
    p_{(m,k),(n,\ell)\vert (o,0)} - p_{(m,k+1),(n,\ell)\vert (o,0)} ={}& p_{(m, k)\vert (o,0)} \mu_\sigma (m) \left( \mathcal{E}_\sigma -1 \right) p_{(n,\ell) \vert (m,k+1 )} \nonumber \\
    {}& - \left( \mathcal{E}_\sigma^\dagger -1 \right) \left [p_{(n,\ell)\vert (m,k+1)} \mu_\sigma (m)\, p_{(m,k) \vert (o,0)}\right] \nonumber \\
    {}& + \left[\mathcal{E}_\sigma^\dagger\left(\mu_\sigma (m)\, p_{(m,k) \vert (o,0)}\right)\right] \left[\left( \mathcal{E}_\sigma^\dagger -1 \right) p_{(n,\ell)\vert (m,k+1)} \right] \nonumber \\
    = {}& p_{(m, k)\vert (o,0)} \mu_\sigma (m) p_{(n,\ell) \vert (\mathcal{E}_\sigma\left[m\right],k+1 )} - p_{(m, k)\vert (o,0)} \mu_\sigma (m) p_{(n,\ell) \vert (m,k+1 )} \nonumber \\
    {}&  - \left( \mathcal{E}_\sigma^\dagger -1 \right) \left [p_{(n,\ell)\vert (m,k+1)} \mu_\sigma (m)\, p_{(m,k) \vert (o,0)}\right] \nonumber \\ 
    {}& + \mu_\sigma (\mathcal{E}_\sigma^\dagger\left[m\right])\, p_{(\mathcal{E}_\sigma^\dagger\left[m\right],k) \vert (o,0)}  p_{(n,\ell)\vert (\mathcal{E}_\sigma^\dagger\left[m\right],k+1)} \nonumber\\
    {}& - \mu_\sigma (\mathcal{E}_\sigma^\dagger\left[m\right])\, p_{(\mathcal{E}_\sigma^\dagger\left[m\right],k) \vert (o,0)} p_{(n,\ell)\vert (m,k+1)}.
\end{align}
The above expanded equation can be simplified by the identity
\begin{align}
\left( \mathcal{E}_\sigma^\dagger -1 \right) \left [p_{(n,\ell)\vert (m,k+1)} \mu_\sigma (m)\, p_{(m,k) \vert (o,0)}\right] ={}& p_{(n,\ell)\vert (\mathcal{E}_\sigma^\dagger\left[m\right],k+1)} \mu_\sigma (\mathcal{E}_\sigma^\dagger\left[m\right])\, p_{(\mathcal{E}_\sigma^\dagger\left[m\right],k) \vert (o,0)} \nonumber \\{}& -p_{(n,\ell)\vert (m,k+1)} \mu_\sigma (m)\, p_{(m,k) \vert (o,0)},
\end{align}
leading to the reverse-time evolutionary equation for the joint conditional probabilities, analogous to Eq.~\eqref{eq:charlie}:
\begin{align}
    p_{(m,k),(n,\ell)\vert (o,0)} - p_{(m,k+1),(n,\ell)\vert (o,0)} = {}& p_{(m, k)\vert (o,0)} \mu_\sigma (m) p_{(n,\ell) \vert (\mathcal{E}_\sigma\left[m\right],k+1 )} - \mu_\sigma (\mathcal{E}_\sigma^\dagger\left[m\right])\, p_{(\mathcal{E}_\sigma^\dagger\left[m\right],k) \vert (o,0)} p_{(n,\ell)\vert (m,k+1)} \nonumber\\
    ={}& \left(\mathcal{E}_{\sigma} - 1 \right) \left[\mu_\sigma (\mathcal{E}_\sigma^\dagger\left[m\right])\, p_{(\mathcal{E}_\sigma^\dagger\left[m\right],k) \vert (o,0)} p_{(n,\ell)\vert (m,k+1)}\right]  \nonumber \\
    ={}&  \left(\mathcal{E}_{\sigma} - 1 \right) \left[\mu_\sigma (\mathcal{E}_\sigma^\dagger\left[m\right]) \frac{p_{(\mathcal{E}_\sigma^\dagger\left[m\right],k) \vert (o,0)}}{p_{(m,k+1) \vert (o,0)}} p_{(n,\ell), (m,k+1)\vert (o, 0)}\right]. 
\end{align} 
Dividing both side by $p_{(n, \ell)\vert (o, 0)}$ leads to:
\begin{equation}
    p_{(m,k)\vert (o,0),(n,\ell)} = p_{(m,k+1)\vert (o,0),(n,\ell)} + \left(\mathcal{E}_{\sigma} - 1 \right) \left[\mu_\sigma (\mathcal{E}_\sigma^\dagger\left[m\right]) \frac{p_{(\mathcal{E}_\sigma^\dagger\left[m\right],k) \vert (o,0)}}{p_{(m,k+1) \vert (o,0)}} p_{ (m,k+1)\vert (o, 0),(n,\ell)}\right],  \label{eq:romeo}
\end{equation} 
which is the reverse-time stochastic process for the discrete-time and discrete-state Markov processes.

\emph{Remark 1.} Both Eqs.~\eqref{eq:david} and \eqref{eq:romeo} can be understood as the forward equation for the reverse-time process. Note that an overall operator $\mathcal{E}_\sigma-1$ is applied to the product of the transition rate and conditional probabilities (cf.~the adjoint operator $\mathcal{E}_\sigma^\dagger-1$ in the forward equation of the forward process, Eqs.~\eqref{eq:dforward2} and \eqref{eq:cforward2}). This indicates that the shift operators for the reverse-time process is adjoint to those in the forward process, meaning the possible transitions are opposite---a state $m$ later in time can only transit to a state $m'_\sigma$ earlier in time, if a transition from $m'_\sigma$ to $m$ is permitted in the forward process. Note, however, the transition rate (or probability) of the reverse-time $m\rightarrow m'_\sigma$ is \emph{not} identical to the transition rate (or probability) of the forward process $m_\sigma'\rightarrow m$. 

\emph{Remark 2.} For the discrete-time system, it is possible to adopt an alternative derivation that is similar to \cite{sohl-dicksteinDeepUnsupervisedLearning2015}, without introducing the formal operator-algebraic derivation in this section. We are interested in deriving the probability $p_{(m'_\sigma, k)\vert (m, k+1), (o,0)}$. Using conditional Bayes formula, we can express
\begin{equation}
    p_{(m'_\sigma, k)\vert (m, k+1), (o,0)} =  \frac{p_{(m, k+1)\vert (m'_\sigma, k), (o,0)} p_{(m'_\sigma, k)\vert (o,0)} }{p_{(m, k+1)\vert (o,0)}}. \label{eq:cBayes}
\end{equation}
By the Markov property of the process, $p_{(m, k+1)\vert (m'_\sigma, k), (o,0)} p_{(m'_\sigma, k)\vert (o,0)}=p_{(m, k+1)\vert (m'_\sigma, k)} $, which is the transition probability $L_{mm'_\sigma}$ (and $\nu_\sigma(m'_\sigma)$ in the context of process Eq.~\eqref{eq:dforward2}). Thus, \eqref{eq:cBayes} prescribes the transition probabilities of the reverse-time evolution, which is described by Eq.~\eqref{eq:romeo}. 

\emph{Remark 3.} A typical technique to derive continuous-time master equations is to consider a discrete-time Markov chain and express the transition probabilities in terms of continuous-time rate constants, $\mu_\sigma (m) = \nu_\sigma (m) \Delta t$, and sending $\Delta t\downarrow 0$. It is clear to see such treatment transforms Eq.~\eqref{eq:romeo} to Eq.~\eqref{eq:david}. 

\section{Generalization of the Theoretical Formulation for Arbitrary Discrete-State Markov Processes} \label{app:generalization}
In this section, we present the generalization for the continuous-time process, defined as the forward Chapman--Kolmogorov Equation \eqref{eq:cforward}. A parallel derivation for the discrete-time case (Eq.~\eqref{eq:dforward2}) is omitted.  

The key idea is to realize that the step operators $\mathcal{E}_{\sigma}$ are simply a way for us to reformulate the transition matrix $L$ in \eqref{eq:cforward}. Let us define rate functions for a general $n$-step transition (those transitions from $m$ to $m'$ such that $\vert m-m'\vert = n$):
\begin{equation}
    \nu_{n,\sigma} (m) := \left \{\begin{array}{ll}
    L^\dagger_{m+\sigma n,m},& \text{if } m+\sigma\cdot n \in \Omega \\
    0, & \text{else.}
    \end{array}\right.
\end{equation}
It suffices to consider $n=1, \ldots M$ for our setup (where the state in the finite-state space $\Omega$ is labeled from 1 to $M$). 
Then, Eq.~\eqref{eq:cforward} can be reformulated as a linear sum of powers of the shift operators
\begin{equation}
    \frac{\text{d}}{\text{d} t} p_{(m,t) \vert (o,0)} = \left[ \left(\mathcal{E}_\sigma^\dagger \right)^n -1 \right] \left[ \nu_{n,\sigma}\left(m\right) p_{(m,t) \vert (o,0)} \right]. 
\end{equation}
We remind the reader that we used the Einstein summation convention to sum over $\sigma \in \left\{+,-\right\}$ and $n\in \left\{0, \ldots M \right\}$ in the above equation. Similarly, the Kolmogorov backward equation is
\begin{equation}
    -\frac{\text{d}}{\text{d} s} p_{(n,t) \vert (m,s)}  =  \nu_{n,\sigma} \left(m\right) \left(\mathcal{E}_\sigma^n -1 \right)  p_{(n,t) \vert (m,s)} 
\end{equation}
As the powers of the step operators are still step operators, the procedure in Appendix \ref{app:reverseTime} can be applied to deliver the reverse-time evolutionary equation:
\begin{equation}
     -\frac{\text{d}}{\text{d} s } p_{(m,s)\vert(n,t),(o,0)} =
      \left [ \nu_{n,\sigma}\left(m'\right) \frac{p_{(m',s) \vert (o,0)}}{p_{(m,s) \vert (o,0)}} p_{(m,s)\vert(n,t),(o,0)} \right], \label{eq:creverse-general}
\end{equation}
where $m'$ is again the pre-image $\left(\mathcal{E}_\sigma^\dagger\right)^n \left[m\right]$ of transition $(n,\sigma)$ mapping $m'$ to $m$ forward in time.

\section{Kramers--Moyal expansion of the reverse-time evolution in discrete-state spaces} \label{app:asymptotics}

In this section, we aim to apply the Kramers--Moyal expansion \cite{vanKampen,gardinerStochasticMethodsHandbook2009,weberMasterEquationsTheory2017} to the both the forward and reverse-time discrete-state process, Eq.~\eqref{eq:creverse} and \eqref{eq:creverse} respectively. Albeit rather standard and straightforward, the analysis on the reversal process requires extra care compared to the usual practices, due to the fact that the (reverse-time) transition rate depends on non-local information (i.e., $\nu_\sigma$ is evaluated at the preimage of viable transitions to the state $m$). 

As usual, we set out the analysis by asserting a large system size $V:=\mathcal{O}\left(X_t\right) \gg 1$. We denote the scaled variables by a tilde, for example, $\tilde{m}:=m/V$, $\tilde{n}:=n/V$, and $\tilde{o}:=o/V$. Similarly, the scaled rate constants $\tilde{\nu}_\sigma(\tilde{m}):= \nu_\sigma(m)/V$. The probabilities $p$'s now relate to a probability density $\rho$ in the continuum (state) limit, $p=\rho \text{d}x = \rho / V$, with $\text{d} x := 1/V \ll 1$. Now for any test function $f$ on the discrete-state space, 
\begin{equation}
    \left(\mathcal{E}_\sigma -1 \right) f(m) = f\left (m + \sigma \cdot 1\right)-f(m) = V \tilde{f}\left(\tilde{m} + \sigma \cdot \frac{1}{V} \right).
\end{equation}
With minor assumptions on the scaled test function $\tilde{f}$, it can be Taylor-expanded, which is the foundation of the Kramers--Moyal expansion, i.e., 
\begin{equation}
    \left(\mathcal{E}_\sigma -1 \right) f(m) = f\left (m + \sigma \cdot 1\right)-f(m) = V \tilde{f}\left(\tilde{m}\right) + \sigma \frac{\text{d}\tilde{f}}{\text{d}\tilde{m}} + \frac{1}{2V} \frac{\text{d}^2\tilde{f}}{\text{d}\tilde{m}^2}.
\end{equation}
Per the constraints pointed out \citet{Pawula67}, one usually asymptotically expands up to $\mathcal{O}\left(1/V^2\right)$ and truncates the higher orders to obtain the Fokker--Planck equation.

For the forward process, the expansion leads to the following It\^o stochastic differential equation 
\begin{equation}
    \text{d} \tilde{X}_t = v\left(\tilde{X}_t\right) \text{d} t + \sqrt{D\left(\tilde{X}_t\right)} \text{d} W_t, \label{eq:KramersMoyalForward}
\end{equation}
where $\tilde{X}_t \approx X_t/V$ is the scaled continuum-state variable, $v(x):= \left[\nu_+(Vx)- \nu_- (V x)\right]/V$ is the drift, and $D\left(x\right) := \left[{\nu}_+(V x) + \nu_- (V x )\right]/(2V^2)$ is the diffusion. The forward solution of the discrete-state system, $p_{(m,t)\vert (o,0)}$, asymptotically converges to the solution of Eq.~\eqref{eq:KramersMoyalForward} in the continuum limit $M\gg 1$. 

Performing the expansion for the reversal process Eq.~\eqref{eq:creverse} requires extra care. The complication from the fact that the transition rates inside the square brackets are non-local and requires asymptotic expansion, and a careful matching of the asymptotic orders is needed. We first rewrite the reverse-time transition rate
\begin{equation}
    \nu_\sigma\left(\mathcal{E}_\sigma^\dagger\left[m\right]\right) \frac{p_{(\mathcal{E}_\sigma^\dagger\left[m\right],s)\vert (o,0)}}{p_{(m,s)\vert(o,0)}}  =  \left[\nu_\sigma\left(\mathcal{E}_\sigma^\dagger\left[m\right]\right)-\nu_\sigma\left(m\right) 
    + \nu_\sigma\left(m\right)  \right]\left[ \frac{p_{(\mathcal{E}_\sigma^\dagger\left[m\right],s)\vert (o,0)}-p_{(m,s)\vert(o,0)}}{p_{(m,s)\vert(o,0)}} + 1\right].   \label{eq:xavier}
\end{equation}
Asymptotically expanding difference terms in both brackets to $\mathcal{O}\left(1/V\right)$:
\begin{subequations}
\begin{align}
    \nu_\sigma\left(\mathcal{E}_\sigma^\dagger\left[m\right]\right)-\nu_\sigma\left(m\right)  ={}& \left(\mathcal{E}_\sigma^\dagger-1\right) \nu_\sigma\left(m\right) = -\sigma \frac{\text{d}\tilde{\nu}_\sigma }{\text{d}\tilde{m}} + \mathcal{O}\left(\frac{1}{V}\right), \\   \frac{p_{(\mathcal{E}_\sigma^\dagger\left[m\right],s)\vert (o,0)}-p_{(m,s)\vert(o,0)}}{p_{(m,s)\vert(o,0)}} = {}& \frac{\left(\mathcal{E}_\sigma^\dagger-1\right) p_{(m,s)\vert(o,0)}}{p_{(m,s)\vert(o,0)}} \nonumber \\ = {}&  -\frac{\sigma}{V} \frac{1}{\rho\left(\tilde{m},s \vert \tilde{o},0\right)} \frac{\text{d} \rho\left(\tilde{m},s \vert \tilde{o},0\right) }{\text{d}\tilde{m}} +\mathcal{O}\left(\frac{1}{V}\right)  \label{eq:yankees}. 
\end{align}
\end{subequations}
A key detail of the analysis is that 
\begin{equation}
    \mathcal{O}\left( \frac{1}{\rho\left(\tilde{m},s \vert \tilde{o},0\right)} \frac{\partial \rho\left(\tilde{m},s \vert \tilde{o},0\right) }{\partial\tilde{m}} \right) =\mathcal{O}\left( \frac{\partial \log\left(\tilde{m}\right)}{\partial\tilde{m}}\right)=\mathcal{O}\left(V\right),
\end{equation}
because the variance of the forward Gaussian diffusion $\rho\left(\tilde{m},s \vert \tilde{o},0\right)$ is of order $\mathcal{O}\left(1/V\right)$. We can now collect the leading ($\mathcal{O}\left(V^1\right)$) asymptotic expression of the transition rates in Eq.~\eqref{eq:xavier}:
\begin{subequations}
\begin{align}
    \nu_\sigma\left(\mathcal{E}_\sigma^\dagger\left[m\right]\right) \frac{p_{(\mathcal{E}_\sigma^\dagger\left[m\right],s)\vert (o,0)}}{p_{(m,s)\vert(o,0)}} = \nu_\sigma\left(m\right) \left[1 - \frac{\sigma}{V} \frac{1}{\rho\left(\tilde{m},s \vert \tilde{o},0\right)} \frac{\text{d} \rho\left(\tilde{m},s \vert \tilde{o},0\right) }{\text{d}\tilde{m}} \right] + \mathcal{O}\left(V^1\right).
\end{align}
\end{subequations}
As such, asymptotically speaking, the transition rates of the reverse-time process are almost identical to the forward process, except for the correction $\left(\sigma/V\right)\partial_{\tilde{m}} \rho\left(\tilde{m}, s\right)$.  
The Kramers--Moyal expansion of the process \eqref{eq:creverse} results in the standard It\^o stochastic differential equation \eqref{eq:KramersMoyalForward}, with the drift $v$ and diffusion $D$ defined as
\begin{subequations}
\begin{align}
    v\left(\tilde{m}\right) ={}&  \sum_{\sigma\in \left\{+,-\right\}} \sigma \cdot \tilde{\nu}_\sigma\left(V\cdot \tilde{m} \right) \left[1 - \frac{\sigma}{V} \frac{1}{\rho\left(\tilde{m},s \vert \tilde{o},0\right)} \frac{\text{d} \rho\left(\tilde{m},s \vert \tilde{o},0\right) }{\text{d}\tilde{m}} \right] \nonumber \\
    ={}&  \left(\sum_{\sigma\in \left\{+,-\right\}} \sigma \cdot \tilde{\nu}_\sigma\left(V\cdot \tilde{m} \right)\right)  - \frac{1}{V} \frac{1}{\rho\left(\tilde{m},s \vert \tilde{o},0\right)} \frac{\text{d} \rho\left(\tilde{m},s \vert \tilde{o},0\right) }{\text{d}\tilde{m}}, \\
    D\left(\tilde{m}\right) ={}&  \frac{1}{2V} \sum_{\sigma\in \left\{+,-\right\}} \tilde{\nu}_\sigma\left(V\cdot \tilde{m} \right). 
\end{align}
\end{subequations}
Note that the functional form of the correction in the drift $v$ is similar to that in \citet{andersonReversetimeDiffusionEquation1982}. Such a correction term is the key target for learning in the score-based generative models \cite{songGenerativeModelingEstimating2020,songImprovedTechniquesTraining2020,songScoreBasedGenerativeModeling2021a,songMaximumLikelihoodTraining2021}.

\emph{Remark 1.} The analogous Stein score function \cite{songGenerativeModelingEstimating2020,songImprovedTechniquesTraining2020,songScoreBasedGenerativeModeling2021a,songMaximumLikelihoodTraining2021} for the discrete-state system can be identified in the above analysis (see Eq.~\eqref{eq:xavier})
\begin{equation}
    s_{\text{dis}, \sigma} \left(m, s\right)\propto \nu_\sigma\left(\mathcal{E}^\dagger_\sigma\left[m\right]\right) \frac{p_{(\mathcal{E}_\sigma^\dagger\left[m\right],s)\vert (o,0)}-p_{(m,s)\vert(o,0)}}{p_{(m,s)\vert(o,0)}} \label{eq:dis-score}
\end{equation}
The asymptotic analysis shows in the dense-grid limit, the above finite-difference formula is asymptotically proportional to the Stein score function for continuous-sate system, $s_\text{cts}\left(m, s\right):=\partial_{\tilde{m}} \log \rho \left(\tilde{m}, s\right)$.  The proportionality came from the constant $1/V$, which is proportional to the variance of the forward solution. Such a proportionality constant is usually normalized in the score-based approaches, see discussion in \citet{songScoreBasedGenerativeModeling2021a}. It is straightforward to show that the score function of Blackout Diffusion Eq.~\eqref{eq:pureDeath} has an interesting functional form
\begin{equation}
    \frac{1}{m+1}\left(\frac{o e^{-t} - m}{1-e^{-t}} - 1 \right).
\end{equation}
We did not directly learn this function in this paper. 

\emph{Remark 2.} The discrete-state score functions Eq.~\eqref{eq:dis-score} depends on the forward solution evaluated at the preimage $m'=\mathcal{E}_\sigma^\dagger\left[m\right]$ of a viable transition to the state $m$. For general Markov transition process (see Appendix \ref{app:generalization}) with $J$ viable transitions into state $m$ (i.e., $J$ non-zero $L^\dagger_{mm'}$, $m\ne m'$), there exists $J$ distinct score functions with the same functional form \eqref{eq:dis-score}.

\section{Deriving the binomial bridge formula} \label{app:binomialBridge}

The Binomial Bridge Formula \eqref{eq:binomialBridge} can be established straightforwardly. For $0\le s \le t$, Eq.~\eqref{eq:fowardSol} states
\begin{align}
    p_{(m,s)\vert (o,0)}={}& \left(\begin{array}{c} o \\ m \end{array}\right) e^{-ms} \left(1-e^{-s}\right)^{o-m} \\
    p_{(n,t)\vert (o,0)}= {}& \left(\begin{array}{c} o \\ n \end{array}\right) e^{-nt} \left(1-e^{-t}\right)^{o-n}.
\end{align}
In addition, the forward propagation from time $s$ to $t$ can be expressed in a similar form, by the Markov property of the process:
\begin{equation}
    p_{(n,t)\vert (m,s)} = \left(\begin{array}{c} m \\ n \end{array}\right) e^{-n\left(t-s\right)} \left(1-e^{-\left(t-s\right)}\right)^{m-n}.
\end{equation}
Applying conditional Bayes formula, 
\begin{align}
    p_{(m,s)\vert (n,t), (o,0)} ={}& \frac{ p_{(n,t)\vert (m,s), (o,0)}  p_{(m,s)\vert (o,0)} }{ p_{(n,t)\vert (o,0)} } = \frac{ p_{(n,t)\vert (m,s)}  p_{(m,s)\vert (o,0)} }{ p_{(n,t)\vert (o,0)} } \nonumber \\
    = {}& \frac{\left(o-n\right)!}{\left(m-n\right)! \left(o-m\right)!} \frac{e^{-ms} \left(1-e^{-s}\right)^{o-m} e^{-n\left(t-s\right)} \left(1-e^{-\left(t-s\right)}\right)^{m-n}}{e^{-nt} \left(1-e^{-t}\right)^{o-n}}\nonumber\\
    = {}& \frac{\left(o-n\right)!}{\left(m-n\right)! \left(o-m\right)!}  \left(\frac{1-e^{-s}}{1-e^{-t}}\right)^{o-m}  \left(\frac{e^{-s}-e^{-t}}{1-e^{-t}}\right)^{m-n},
\end{align}
where we used the Markov property to establish the first equality. By defining $r:=\left(e^{-s}-e^{-t}\right)/\left(1-e^{-t}\right)$, the above equation can be succinctly expressed as
\begin{equation}
    p_{(m,s)\vert (n,t), (o,0)} =\frac{\left(o-n\right)!}{\left(m-n\right)! \left(o-m\right)!}  \left(1-r\right)^{o-m} r^{m-n},
\end{equation}
which is the probability mass function of a binomial distribution. 

\emph{Remark (the ``physicists' solution'').} The Binomial Bridge Formula can be understood intuitively by the following argument. Consider $o$ radioactive particles going through a $\beta$ decay process. We are asked to infer how many are radioactive at time $s$, given that there are $n$ still active at time $t$. Another way to frame the condition is that at time $t$, there are $o-n$ already decayed. For any one of these particles, the conditional probability that it already decayed at time $s$ conditioned on that it decayed at time $t$ is $(1-e^{-s})/(1-e^{-t})$; equivalently, the conditional probability that it remained radioactive at time $s$ conditioned on that it decayed at time $t$ is $1-(1-e^{-s})/(1-e^{-t}) = (e^{-s}-e^{-t})/(1-e^{-t})$. Because each particle goes through an independent process, among those $o-n$ which decayed at time $t$, $\text{Binom}\left(o-n, \left(e^{-s}-e^{-t}\right)/\left(1-e^{-t}\right) \right)$ remained radioactive at time $s$. Adding $n$ which has not decayed by time $t$, the total remaining population at time $s$ is thus $n + \text{Binom}\left(o-n, \left(1-e^{-s}\right)/\left(1-e^{-t}\right) \right)$ remaining.

\section{Observation times} \label{app:schedule}

The observation times $\left\{t_k\right\}_{k=1}^T$ specify when the realizations of the forward stochastic processes are generated. The choice of $\left\{t_k\right\}_{k=1}^T$, analogous to the noise schedules in Gaussian diffusion models \cite{sohl-dicksteinDeepUnsupervisedLearning2015,songScoreBasedGenerativeModeling2021a}, is crucial to the quality of the trained generative model. Here, we present a heuristic way to design the observation times. 

\begin{figure*}[h!]
    \centering
    \includegraphics[ width=1.0\textwidth]{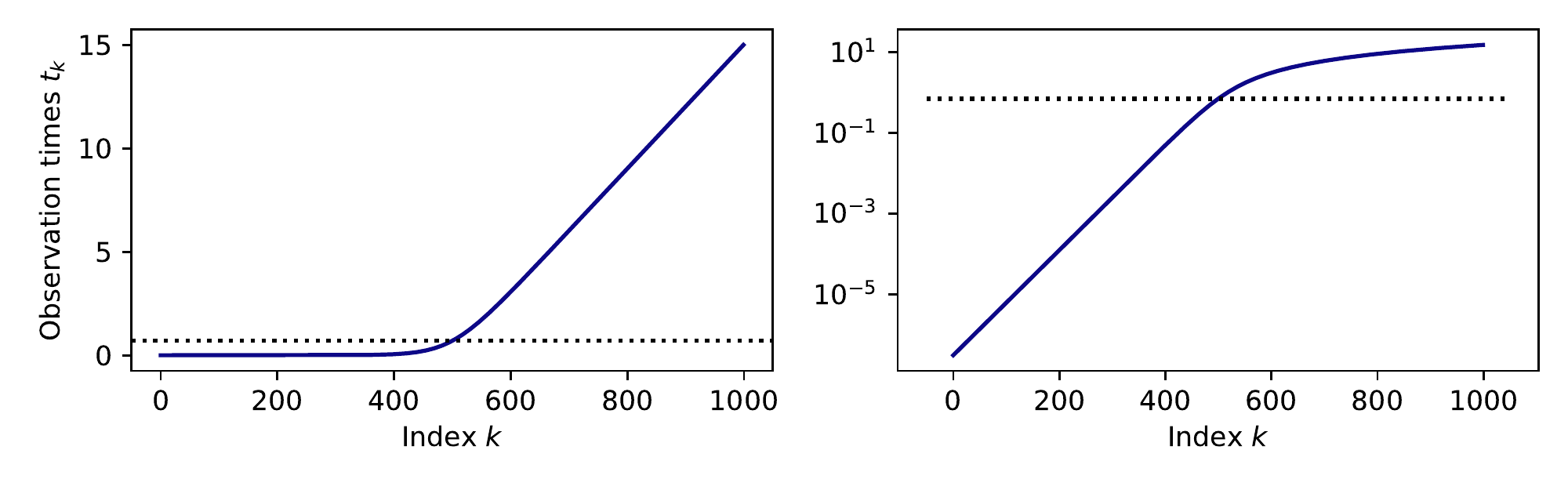}
    \caption{The observation times defined by Eq.~\eqref{eq:observationTimes}, in (left panel) linear and (right panel) logarithmic scales. The dashed lines label the time $t=\log 2$, when the images are maximally noisified.} 
    \label{fig:observationTimes}
\end{figure*}

Our proposition is to adopt the Fisher information (FI) of the forward process as a guidance. The main idea is that we will use more observation times when the forward process \eqref{eq:cforward2} has higher FI. Recall that for the pure-death process \eqref{eq:pureDeath}, suppose the color pixel at $t=0$ is at a state $o\in \Omega$, at time $t$ the solution to the forward process is $\text{Binom}\left(o, q(t)\right)$, where the parameter $q(t)=\exp\left(-t\right)$. As such, the FI of parameter $q$ can be expressed
\begin{equation}
    \text{FI} (q) = \frac{o}{q\left(1-q\right) }. 
\end{equation}
Importantly, the time-dependent part of the $\text{FI}(q)$ is identical for all initial configuration; the implicit dependence comes from $q(t)=\exp\left(-t\right)$. Next, let us define a probability distribution $\phi$ of the observation times in infinitely-many observation limit ($T\rightarrow \infty$), such that the corresponding density $\text{d} \phi$ is proportional to the time-dependent part of the FI:
\begin{equation}
    \text{d} \phi \left(q\right) \propto \frac{1}{q\left(1-q\right) }.
\end{equation}
With a change of variable from $q$ to $t$ by $q=e^{-t}$, the density function $\phi$ on the time domain is
\begin{equation}
    \text{d} \phi\left(t \right) \propto \frac{e^{-t}}{e^{-t} \left(1-e^{-t}\right)}.  \label{eq:quebec}
\end{equation}
Note that the function $\text{d}\phi$ is symmetric about $t=\log(2)$, at which time the system is maximally noisy. 
Next, we use the inverse transform sampling to generate discrete observation times. The first step is to express the cumulative distribution function, obtained by integrating \eqref{eq:quebec}
\begin{equation}
    \text{CDF} (t) \propto \text{Logit}\left(e^{-t}\right) + const.
\end{equation}
Note the apparent singularity as $t\downarrow 0$ and $t\uparrow \infty$. We bypass the singularity by a symmetry argument, that we would like to put exactly half of the discretization times before and after $\log(2)$, and the fact that the final time $t_T=15<\infty $ will be chosen. With the symmetry argument, the first finite time $t_1$ is determined to be $-\log\left(1-e^{t_T}\right)$. Then, we uniformly choose in between $\text{CDF} (t_1)$ and $\text{CDF} (T)$ for the rest $t_k$'s, i.e.,
\begin{equation}
    t_k = -\log \left[ \sigma \left(\text{Logit}\left(1-e^{-t_T}\right) + \frac{k-1}{T-1} \left[\text{Logit}\left(e^{-t_T}\right) -\text{Logit}\left(1-e^{-t_T}\right)\right] \right) \right], \quad k=1, 2, \ldots T, \label{eq:observationTimes}
\end{equation}
where $\sigma(x)\equiv \text{Logit}^{-1}(x) :=1/\left(1+e^{-x}\right)$ is the sigmoid function. 
Figure \ref{fig:observationTimes} shows the dependence of $t_k$ on $k$. 

It is worth mentioning that we also experimented other heuristics, including replacing the Fisher Information by the entropy of the process, entropic production rate, variance, Kullback--Leibler divergence between observation times, signal-to-noise ratio metrics such as the coefficients of variation, and uniform temporal grid. Among all the heuristic methods we experimented, the Fisher Information performed significantly better. 

\section{Likelihood functions} \label{app:likelihood}

In this section, we derive the loss function from the formal likelihood function of continuous-time and discrete-state system. It suffices to consider one transition (say, $m\rightarrow m'$) with aground-truth transition rate $\lambda(t)$. In the limit of $\delta t \ll 1$, the probability of the state transition to $m'$ is then $\lambda(t) \Delta t$ \cite{vanKampen,gardinerStochasticMethodsHandbook2009}, and the probability of no transition event occurred is $1-\lambda(t) \Delta t$. More-than-one transition events can be ignored because they occur at higher order of $\mathcal{O}\left(\Delta t^2\right)$. Let us denote the model-predicted transition rate by $\kappa_\theta(t)$ which depends on model parameters $\theta$. The corresponding model predicted probabilities are $\kappa_\theta(t) \Delta t$  and $1-\kappa_\theta(t) \Delta t$. Now, the support of the viable state is only $\left\{m,m'\right\}$, so the instantaneous (at time $t$) negative log-likelihood (NNL) can be easily formulated
\begin{equation}
    -\text{NLL}(t) = \lambda(t) \Delta t \log \frac{\lambda(t) \Delta t }{\kappa_\theta(t) \Delta t} + \left(1-\lambda(t) \Delta t\right) \log \frac{1-\lambda(t) \Delta t}{1-\kappa_\theta(t) \Delta t}
\end{equation}
which is the Kullback--Leibler divergence between the two Bernoulli distributions induced by the ground-truth and model-predicted processes at time $t$. Neglecting $\theta$-independent terms, and asymptotically expanding the second term on the RHS
\begin{equation}
    - \left(1-\lambda(t) \Delta t\right) \log \left(1-\kappa_\theta(t) \Delta t)\right) =  \kappa_\theta(t) \Delta t + \mathcal{O}\left(\Delta t^2\right), 
\end{equation}
we arrive at
\begin{equation}
    -\text{NLL}(t) = \Delta t \left( \kappa_\theta(t) -\lambda(t)  \log \kappa_\theta (t) \right) + (const. \text{ of } \theta).
\end{equation}
Note that in the reversal process, both the ground-truth reversal rate (prescribed by Eq.~\eqref{eq:creverse}) and the model-predicted rate $\kappa_\theta$ are time-dependent, and we want to learn the whole process, formally $0\le t <\infty$. Let us first consider the full negative log-likelihood in the continuum-time limit (i.e., infinitely many observation times) by integrating the above expression:
\begin{equation}
    \text{Full NLL} = \int \left( \kappa_\theta \left(\tau \right) -\lambda  \left(\tau \right)\log \kappa_\theta \left(\tau \right) \right) \text{d} \tau + (const. \text{ of } \theta).
\end{equation}
In practice, the times are sampled in a Monte Carlo fashion, such that only one observation time is drawn for each drawn training image. Recall that we devised a sampling distribution, Eq.\eqref{eq:quebec}, based on a heuristic argument about Fisher information of the process in Appendix \ref{app:schedule}. Similar to an importance sampler, each sample is reweighted by the probability density $\phi(t)$:
\begin{equation}
    \text{Full NLL} = \int \phi\left(\tau\right) \left[\frac{1}{\phi\left(\tau \right)}\left( \kappa_\theta \left(\tau \right) -\lambda  \left(\tau \right)\log \kappa_\theta \left(\tau \right) \right)\right] \text{d} \tau + (const. \text{ of } \theta).
\end{equation}
This brought us to the loss function as
\begin{equation}
    l =  \frac{\Delta t}{\phi\left(\tau \right)} \left(\kappa_\theta -\lambda \log \kappa_\theta\right),
\end{equation}
and by minimizing $l$ we can achieve maximization of the likelihood. For the Blackout Diffusion process, the reverse-time transition rate 
\begin{equation}
    \lambda = \left(X_0-X_{t_{k+1}}\right) \frac{\left(t_k-t_{k-1}\right) e^{-{t_{k}}}}{1-e^{-{t_{k}}}},
\end{equation}
and as the neural net predicts also the difference $y=o-m$,  
\begin{equation}
    \kappa_\theta  = y \times \frac{e^{-t_{k}}}{1-e^{-t_{k}}},
\end{equation}
which leads to the expression \eqref{eq:lossInstantaneous}, after neglecting another $\theta$-independent constant.  

We next derive finite-time corrected loss function, Eq.~\eqref{eq:lossFinite}. Note that the model predicted $\kappa_\theta$ can only be evaluated at \emph{a priori} defined observation times, $t_k$'s. Using the (constant) rates predicted at time $t_{k}$ to evolve the system back to $t_{k-1}$ is identical to the $\tau$-leaping algorithm (\citet{tauLeaping21}; it is analogous to the Euler--Maruyama integration scheme for continuous-state systems). Without taking time-dependent rate constants into account, the constant-rate approach results in $X_{t_{k}} \vert \left(X_0, X_{t_{k+1}}\right) - X_{t_k} \sim \text{Poisson}\left(\kappa_\theta\right)$, and forms only an approximation to the true binomial distribution \ref{app:binomialBridge}. In this case, we can still formulate the exact likelihood: assuming that at time $t_{k+1}$
\begin{equation}
     -\log L = \sum_{j=m}^{M} \text{PMF}_\text{GT} \left(m\right) \log \frac{\text{PMF}_\text{GT} \left(m\right)}{\text{PMF}_\theta \left(m\right)},
\end{equation}
where $\text{PMF}_\text{GT}$ is the ground-truth probability mass function of the binomial Eq.~\eqref{eq:binomialBridge}, and $\text{PMF}_\theta$ is the model-predicted probability mass function, i.e., $\text{Poisson}\left(\kappa_\theta\right)$. Inserting the expressions, we obtain
\begin{align}
     -\log L ={}& - \sum_{j=m}^{M} \text{PMF}_\text{GT} \left(m\right) \left[m \log \kappa_{\theta} - \kappa_\theta \right] + (const. \text{ of } \theta) \nonumber \\
        ={}& \kappa_\theta - \left[\sum_{j=m}^{M} m \text{PMF}_\text{GT} \left(m\right) \right] \log \kappa_{\theta}+ (const. \text{ of } \theta) \nonumber \\
        ={}& \kappa_\theta - \left(o-m\right) \frac{e^{-t_{k-1}}-e^{-t_{k}}}{1-e^{-t_{k}}} \log \kappa_{\theta} + (const. \text{ of } \theta).
\end{align}
Since the neural net was used to predict $o-m$, we obtain \eqref{eq:lossFinite}, after discounting the distribution $1/\phi$. 

\section{Algorithms for general continuous-time discrete-state Markov processes} \label{app:generalAlgo}

With a little more effort, it is possible to prescribe algorithms for general continuous-time discrete-state Markov processes. Let us assume that at each discrete state $m$ in $\Omega$, there are $R$ potential transitions $m\rightarrow m'$. We will use $r=1\ldots R$ to denote the type of transition event, $m'_r$ as the pre-image of the transition event (i.e., $m'_r\rightarrow m$ via type $r$ transition), and $\nu_r(m)$ as the forward transition rate. We will assume that the forward solution $p_{(m,t)\vert(o,0)}$ is already solved and provided (see discussion in Sec.~\ref{sec:discussion}).

By Eq.~\eqref{eq:creverse}, we can construct the reversal transition rate $m\rightarrow m'_r$:
\begin{equation}
    \lambda_r\left(m\right) := \nu\left(m'_r\right)\frac{p_{(m'_r,t)\vert(o,0)}}{p_{(m,t)\vert(o,0)}}, \label{eq:reverseRate}
\end{equation}
which will be the learning target. Next, assume that the neural network, which is used to approximate the reversal transition rates, is augmented to account for the index $r$ (a simple but most likely not optimal implementation route is to use $R$ independent neural nets like NCSN++). A parallel derivation to that of the instantaneous likelihood in Appendix \ref{app:likelihood} leads to 
\begin{equation}
    -\log L = \sum_{r=1}^R \Delta t \left( \kappa_{r;\theta} -\lambda_r \log \kappa_{r;\theta} \right) + (const. \text{ of } \theta).
\end{equation}
The above equation agrees \citet{campbell2022}. As our construct of the reverse-time stochastic process is exact, this indicates that the variational bounds of \citet{campbell2022} in the continuum-time limit is tight.
As the binomial bridge formula no long holds in general, we use this instantaneous version of the likelihood function (cf.~\eqref{eq:lossInstantaneous}) without the finite-time correction (cf.~\eqref{eq:lossFinite}). For training, we can either use the direct sum, or a Monte Carlo scheme to sample a particular $r$ for each sample. We propose a sampling approach in Algorithm \ref{alg:Gtraining}. Algorithm \ref{alg:Gtraining} and \ref{alg:Ginference} prescribes our proposed training and inference procedures. For inference, the binomial bridge formula no longer holds for the general cases. We thus fold back to the $\tau$-leaping algorithm \cite{tauLeaping21} in Algorithm \ref{alg:Ginference}. We remark that Algorithms \ref{alg:training} and \ref{alg:inference} can be considered as special cases of \ref{alg:Gtraining} and \ref{alg:Ginference}. Our preliminary analysis [data not shown] shows that it is possible to learn a birth-and-death process ($R=2$; \cite{vanKampen}) for the CIFAR-10 dataset. As the aim of this paper is to present the theoretical foundation and to establish the feasibility, we focus on Blackout Diffusion and leave the more complex model to a future study.

\section{Binarized MNIST dataset}
\begin{figure*}[t!]
    \centering
    \includegraphics[width=1.0\textwidth]{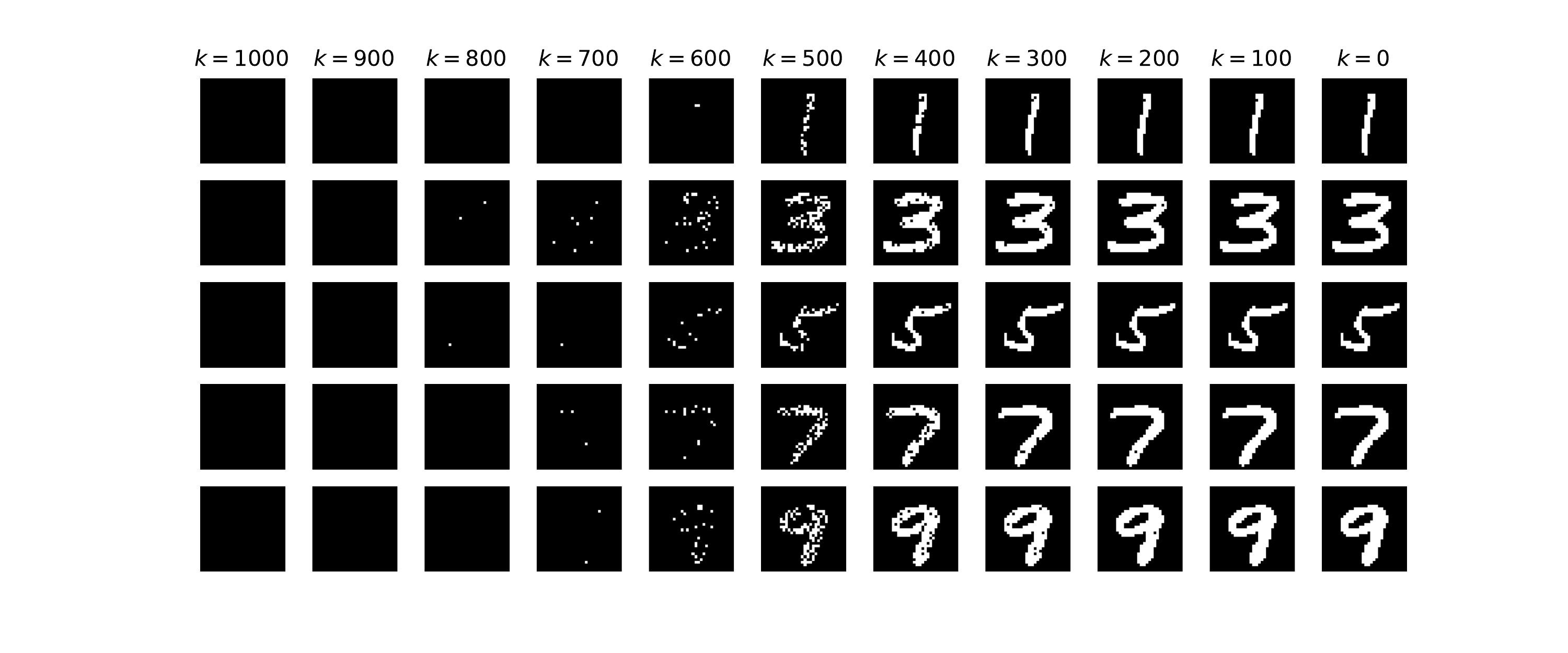}
    \caption{Generation of binarized MNIST samples.}
    \label{fig:mnist-generation}
\end{figure*}

\begin{figure*}[t!]
    \centering
    \includegraphics[width=1.0\textwidth]{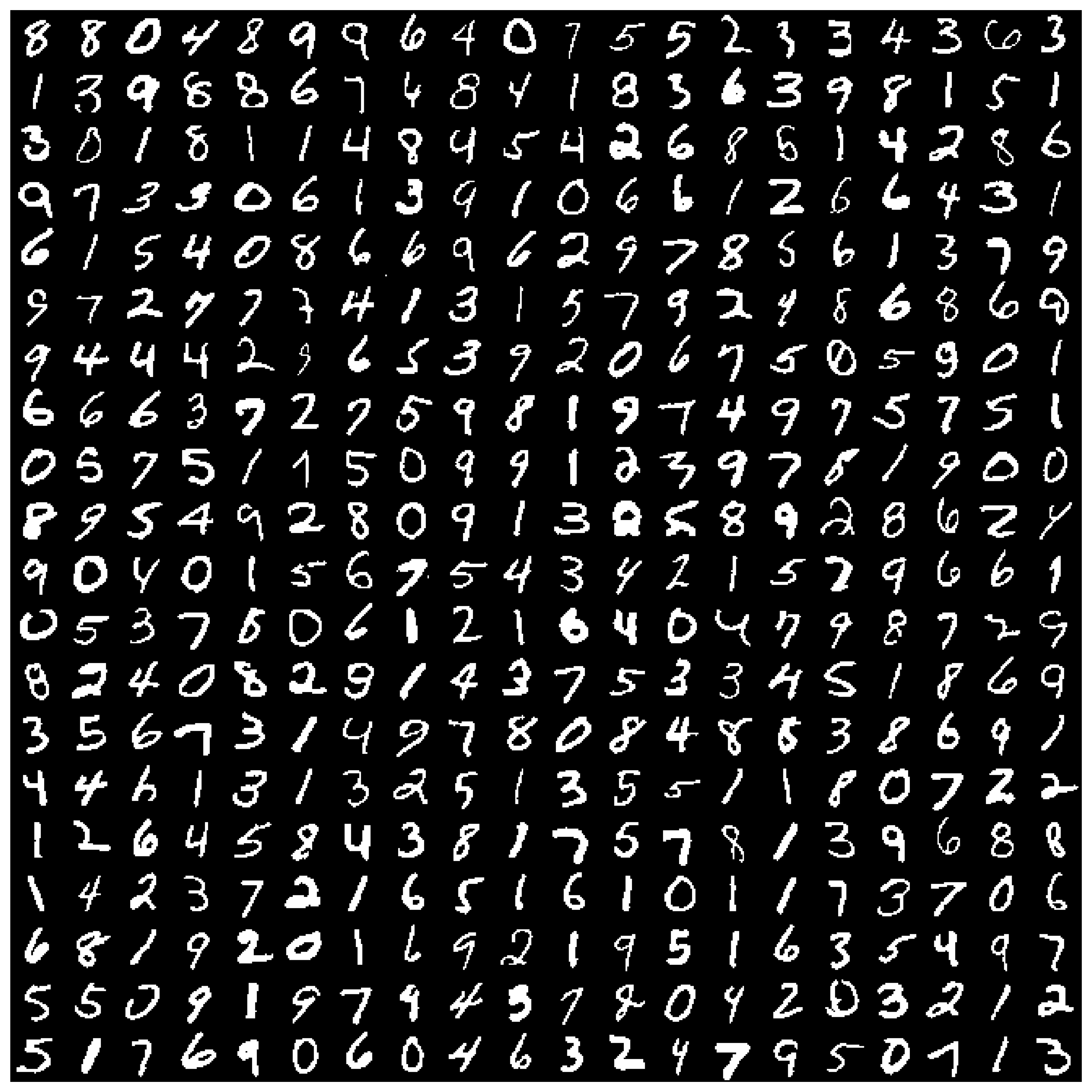}
    \caption{Four hundred binarized MNIST samples generated by Blackout Diffusion.}
    \label{fig:mnist-grid}
\end{figure*}

We performed a parallel analysis on training the Blackout Diffusion Model on a binarized MNIST dataset. We first resized the MNIST samples to $32\times 32$, as the architecture provided in \citet{scorenet} does not apply to $28\times 28$ images. We binarized the dataset by a threshold 127.5. The binarized dataset thus has a highly discrete state space, $\Omega=\left\{0,1\right\}$. We used the infinitesimal loss function \ref{eq:lossInstantaneous}, batch size $256$, and trained the network for 250K iterations. We used the Poissonian scheme for generating the dataset. Figure \eqref{fig:mnist-generation} shows the generation and Fig.~\ref{fig:mnist-grid} showcases 400 generated samples. Using 60,000 generated samples, the FID to the training dataset is $0.023$.

\section{Celeb-A 64x64 dataset}
\begin{figure*}[t!]
    \centering
    \includegraphics[width=1.0\textwidth]{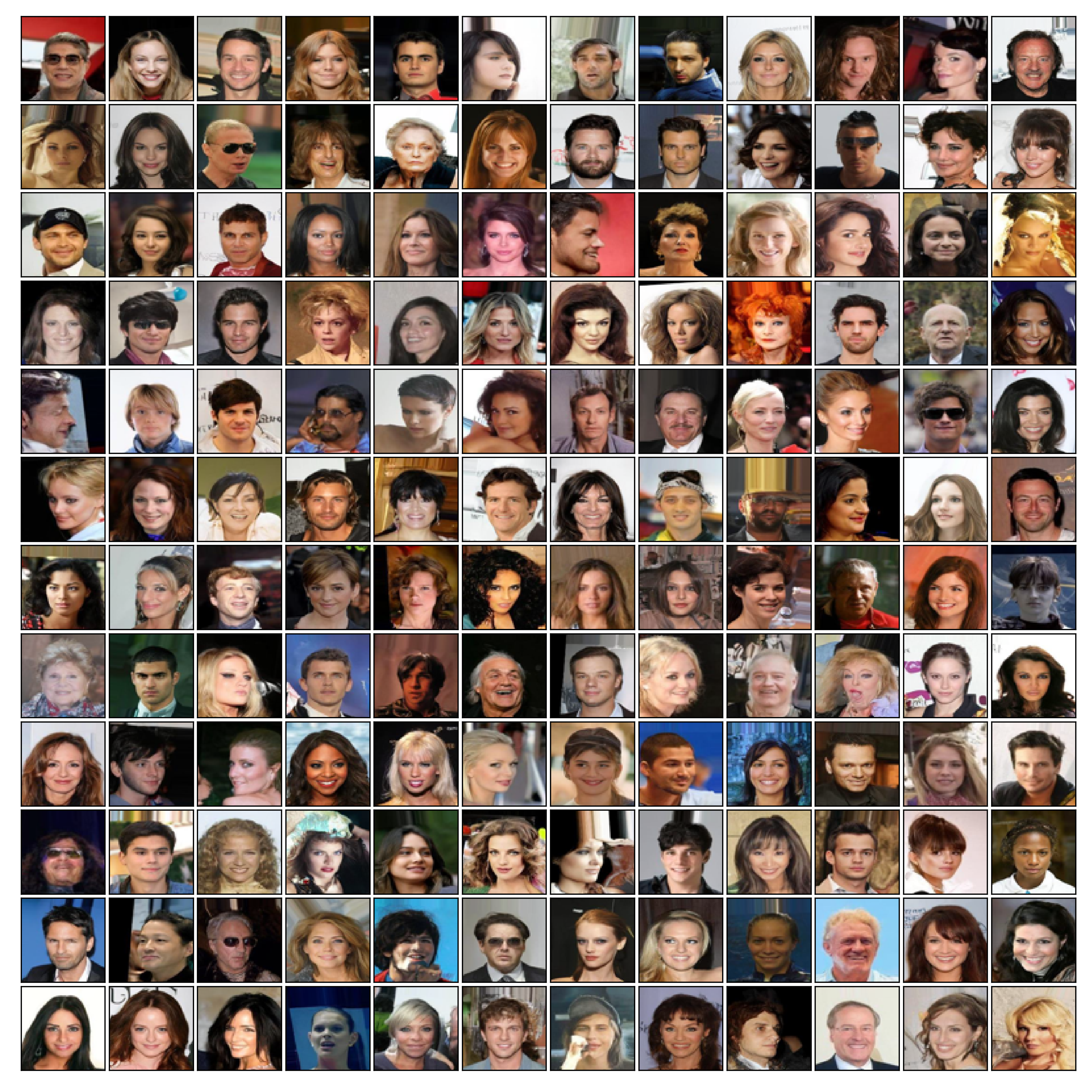}
    \caption{144 samples generated by Blackout Diffusion trained on the Celeb-A dataset by the instantaneous loss function Eq.~\eqref{eq:lossInstantaneous} and generated by $\tau$-leaping (Poisson random number) during inference (FID=3.22).}
    \label{fig:celebA}
\end{figure*}

We conducted a similar study, training our model on the CelebA 64x64 dataset.  We used the infinitesimal loss function \ref{eq:lossInstantaneous} which achieved an FID of $3.22$ when comparing 50,000 generated samples to the training dataset, demonstrating the high fidelity of the model's output. The first 144 generated images can be seen in Fig.~\ref{fig:celebA}.

\begin{algorithm}[tb]
   \caption{Training for general Markov processes}
   \label{alg:Gtraining}
\begin{algorithmic}
   \STATE {\bfseries Input:} Forward solution $p_{(\mathbf{X}_t,t)\vert (\mathbf{X}_0,0)}$, reverse rate functions $\left\{\lambda_r \left(\mathbf{X}\right)\right\}_{r=1}^R$
   \REPEAT
   \STATE $\mathbf{X}_0 \leftarrow \mathbf{x}$, drawn from the training set
   \STATE Draw an index $k$ from $\left\{1, \ldots T\right\}$ uniformly
   \STATE Draw an index $r$ from $\left\{1, \ldots R\right\}$ uniformly
   \STATE $\mathbf{X}_t \sim p_{(\mathbf{X}_t,t)\vert (\mathbf{X}_0,0)}$ (element-wise)
   \STATE $\mathbf{y}_r \leftarrow \mathbf{NN}_\theta \left(\mathbf{X}_{t_k}, k, r\right)$
   \STATE $ w_k \leftarrow \left(t_k - t_{k-1}\right) \left(1-e^{-t_k}\right) $  (Using Eq.~\eqref{eq:lossInstantaneous}) 
   \STATE $l \leftarrow w_k \times \text{mean} \left\{ \left(t_k - t_{k-1}\right)\left[\mathbf{y}_r - \lambda_r \left(\mathbf{X}_{t_k}\right) \log \mathbf{y}_r \right]\right\}$
   \STATE Take a gradient step on $\nabla_\theta l$
   \UNTIL{Converged}
\end{algorithmic}
\end{algorithm}

\begin{algorithm}[tb]
   \caption{Generating images by $\tau$-leaping}
   \label{alg:Ginference}
\begin{algorithmic}
   \STATE Initiate an all-black image $\mathbf{X}_{t_{T}}=0$ 
   \FOR{$k=T$ {\bfseries to} $1$}
   \FOR{$r=1$ {\bfseries to} $R$}
   \STATE $\mathbf{\lambda}_{r;\theta} \leftarrow   \mathbf{NN}_\theta \left(\mathbf{X}_{t_k}, k, r\right),$ 
   \STATE $\mathbf{n}_{r;\theta} \sim \text{Poisson}\left(\mathbf{y}_\theta\right)$ (element-wise)
   \ENDFOR
   \FOR{Each component $i$ of $\mathbf{X}_{t_k}$}
   \FOR{$r=1$ {\bfseries to} $R$}
   \STATE Perform $\left(\mathbf{n}_{r,\theta}\right)_i$ type-$r$ reversal transition
   \ENDFOR
   \ENDFOR
   \ENDFOR
\end{algorithmic}
\end{algorithm}

\newpage

\begin{table}[]
\caption{Continuous-state methods vs Blackout}
\label{tab:cifar10_SOTA}
\vskip 0.15in
\begin{center}
\begin{small}
\begin{sc}
\begin{tabular}{@{}ccccc@{}}
\toprule
Methods on CIFAR10 & Continuous?  & \multicolumn{1}{l}{Sampling steps} & \multicolumn{1}{l}{FID (↓)} &  \begin{tabular}[c]{@{}c@{}}Training \\ iterations\end{tabular} \\ \midrule
Improved++ \cite{nicholImprovedDenoisingDiffusion2021}         &       $\checkmark$                & 1k                                 & 3.29                        & 500k                                           \\
Improved++ \cite{nicholImprovedDenoisingDiffusion2021}              &       $\checkmark$             & 4k                                 & 2.90                        & 500k                                           \\
SDE VE (Deep NCSN++) \cite{songScoreBasedGenerativeModeling2021a}    &      $\checkmark$             & 2k (1k corrector)                  & 2.2                         & 1.3M                                           \\
SDE VE (NCSN++)\cite{songScoreBasedGenerativeModeling2021a}          &      $\checkmark$            & 2k (1k corrector)                  & 2.38                        & 1.3M                                           \\
EDM-G++ \cite{sota}           &     $\checkmark$             & 1k                                 & \textbf{1.77}               & Pre-trained model                              \\ \midrule
Blackout (OURS)                  &      $\times$           & 1k                                 & 4.77                        & \textbf{300k}                                  \\ \bottomrule
\end{tabular}
\end{sc}
\end{small}
\end{center}
\vskip -0.1in
\end{table}

\begin{figure*}[t!]
    \centering
    \includegraphics[width=1.0\textwidth]{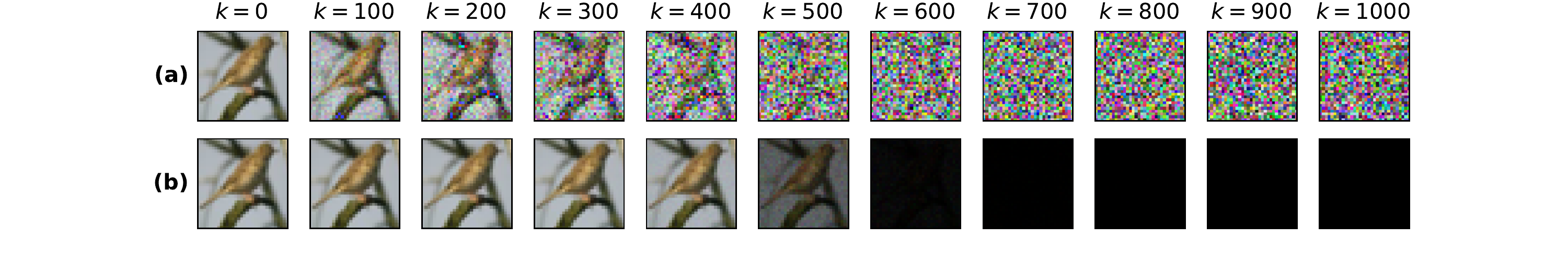}
    \caption{Figure \ref{fig:imageDegradation} without an adjusted colormap.}
    \label{fig:imageDegradation_noScaling}
\end{figure*}

\begin{figure*}[t!]
    \centering
    \includegraphics[width=1.0\textwidth]{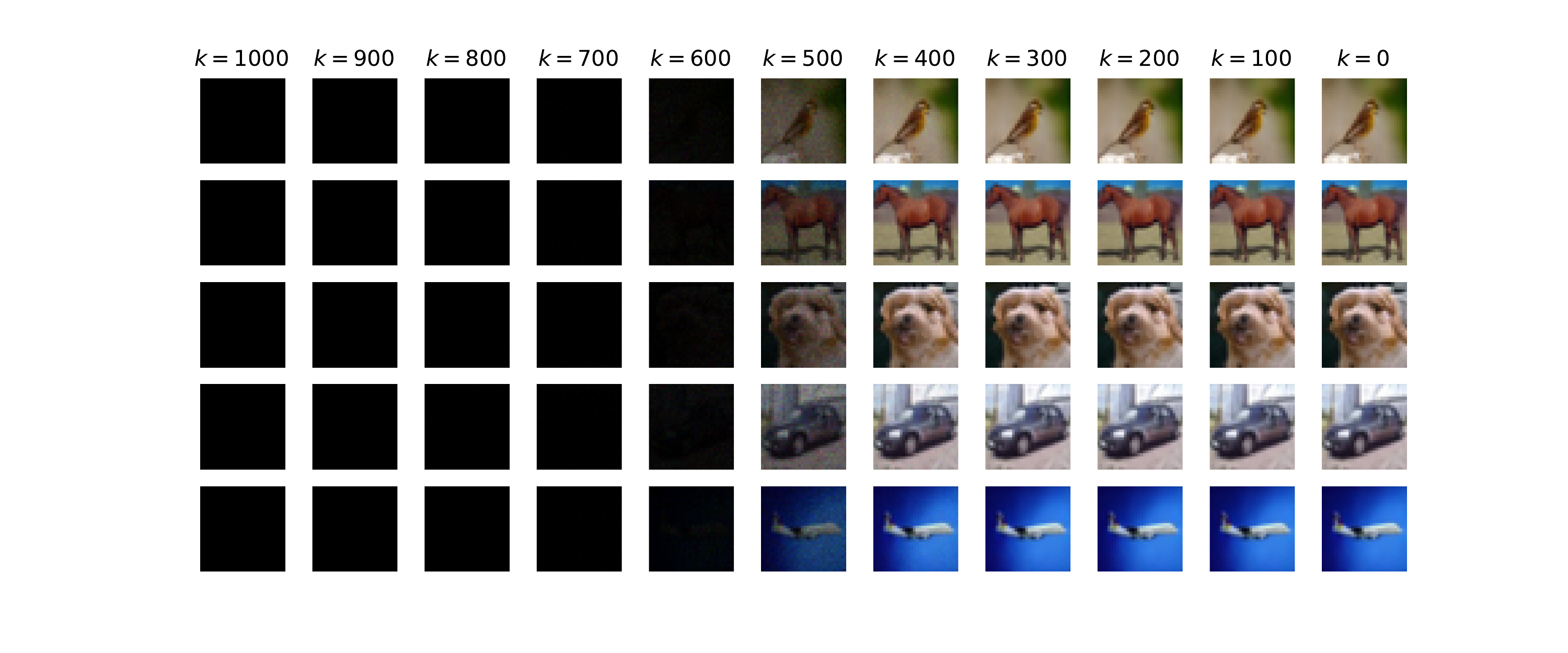}
    \caption{Figure \ref{fig:imageGeneration} without an adjusted colormap.}
    \label{fig:imageGeneration_noScaling}
\end{figure*}

\begin{figure*}[t!]
    \centering
    \includegraphics[width=1.0\textwidth]{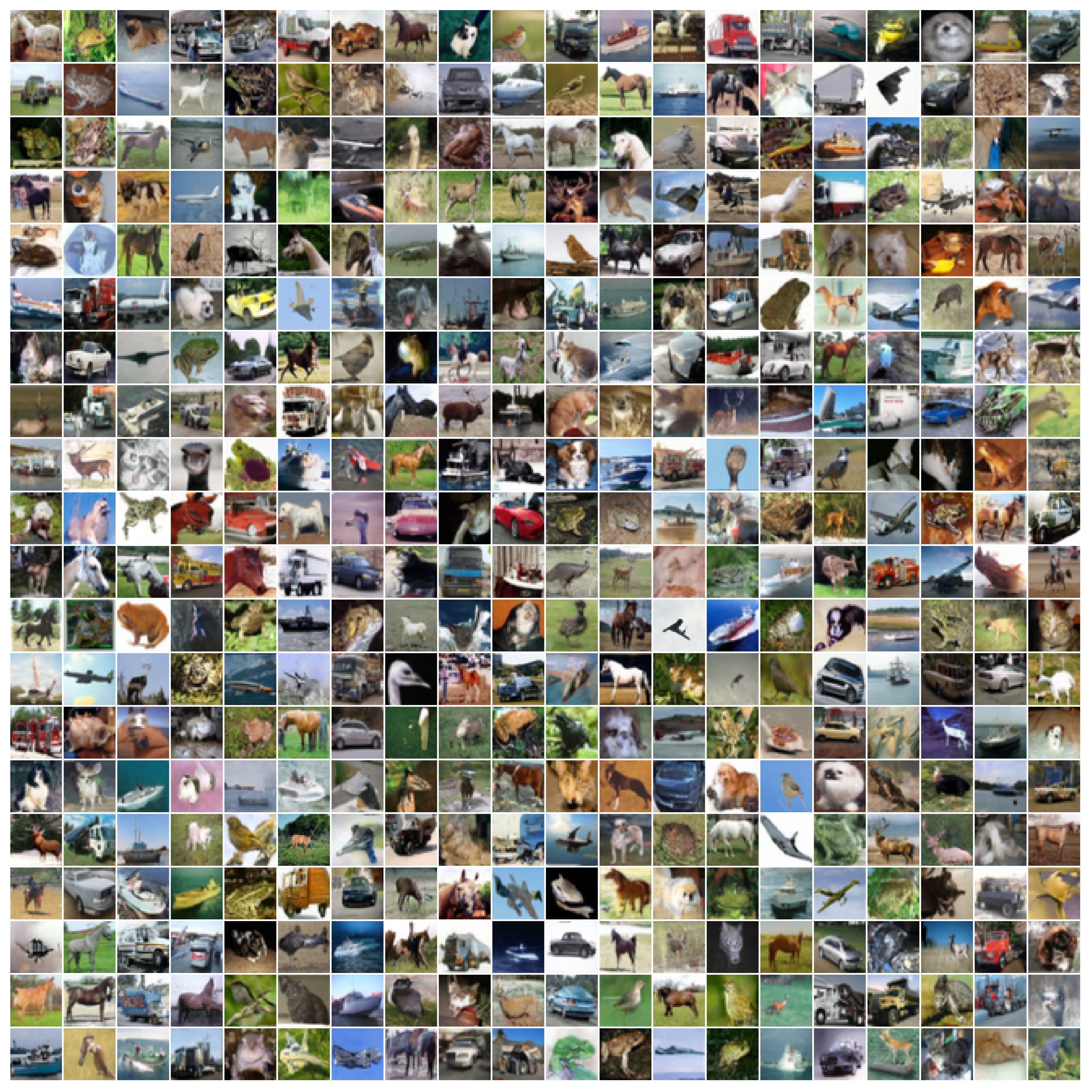}
    \caption{Four  hundred samples generated by Blackout Diffusion trained by the finite-time loss function Eq.~\eqref{eq:lossFinite} and generated by the binomial bridge formula Eq.~\eqref{eq:binomialBridge} during inference.}
    \label{fig:grid-f-b}
\end{figure*}

\begin{figure*}[t!]
    \centering
    \includegraphics[width=1.0\textwidth]{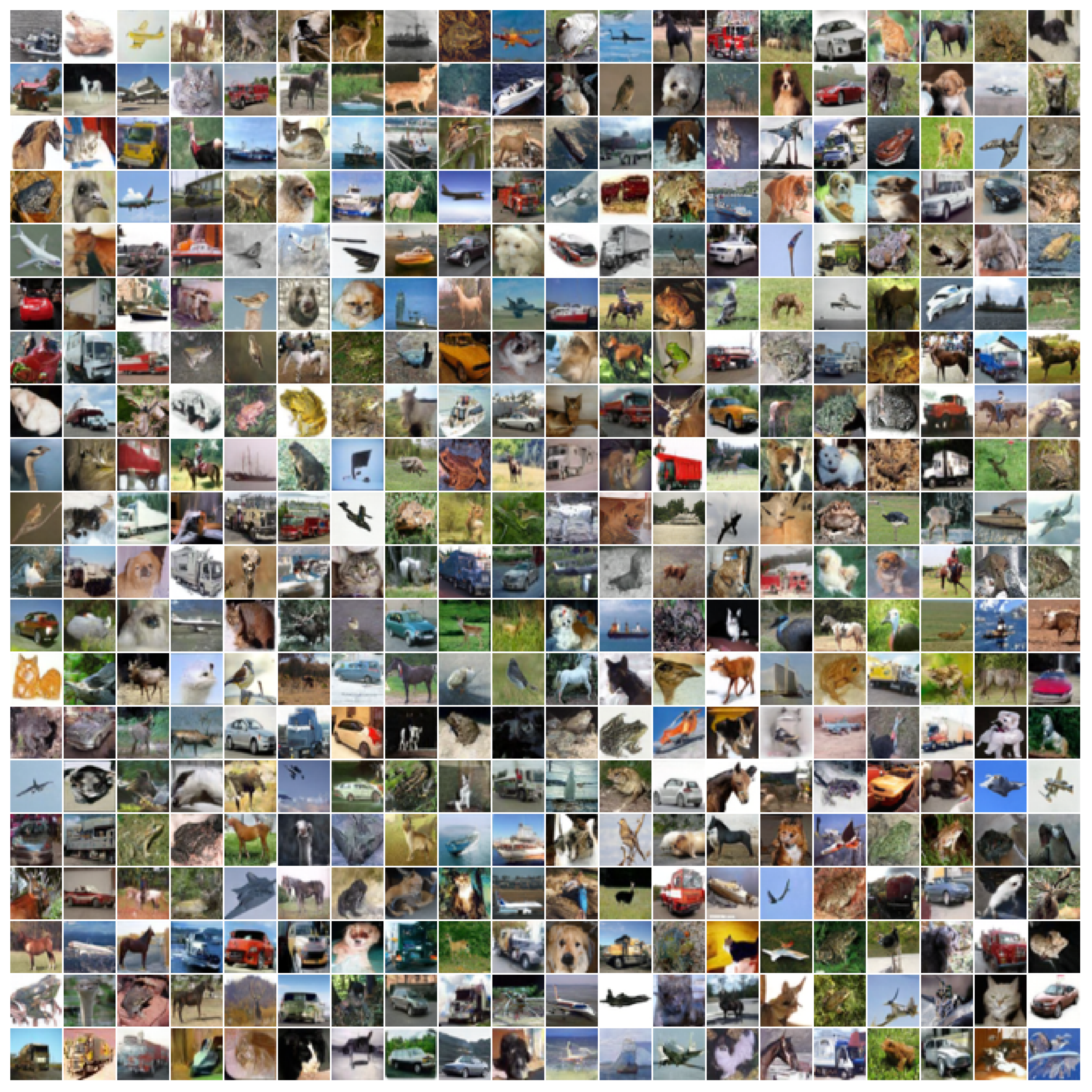}
    \caption{Four  hundred samples generated by Blackout Diffusion trained by the finite-time loss function Eq.~\eqref{eq:lossFinite} and generated by $\tau$-leaping (Poisson random number) during inference.}
    \label{fig:grid-f-p}
\end{figure*}

\begin{figure*}[t!]
    \centering
    \includegraphics[width=1.0\textwidth]{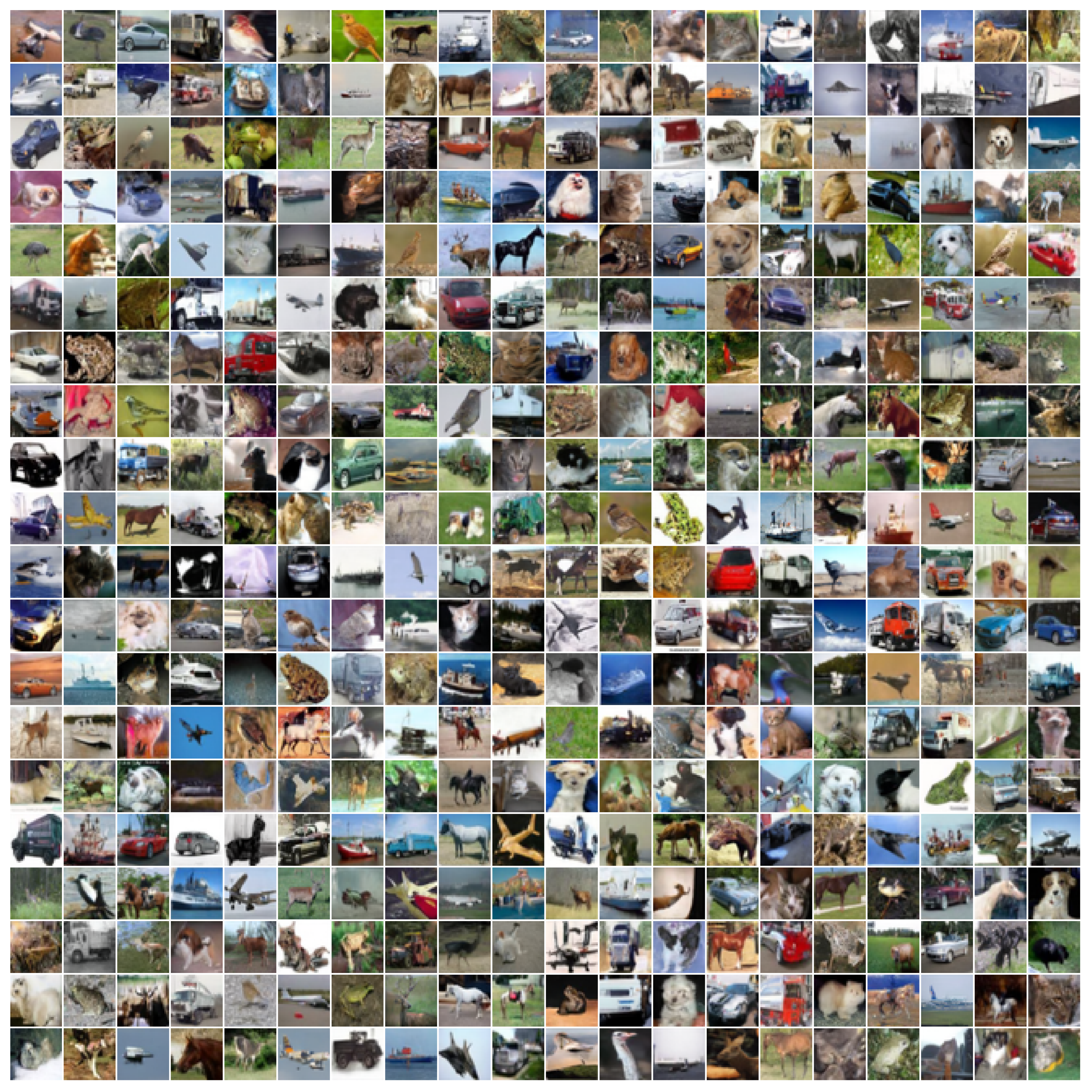}
    \caption{Four hundred samples generated by Blackout Diffusion trained by the instantaneous loss function Eq.~\eqref{eq:lossInstantaneous} and generated by the binomial bridge formula Eq.~\eqref{eq:binomialBridge} during inference.}
    \label{fig:grid-i-b}
\end{figure*}

\begin{figure*}[t!]
    \centering
    \includegraphics[width=1.0\textwidth]{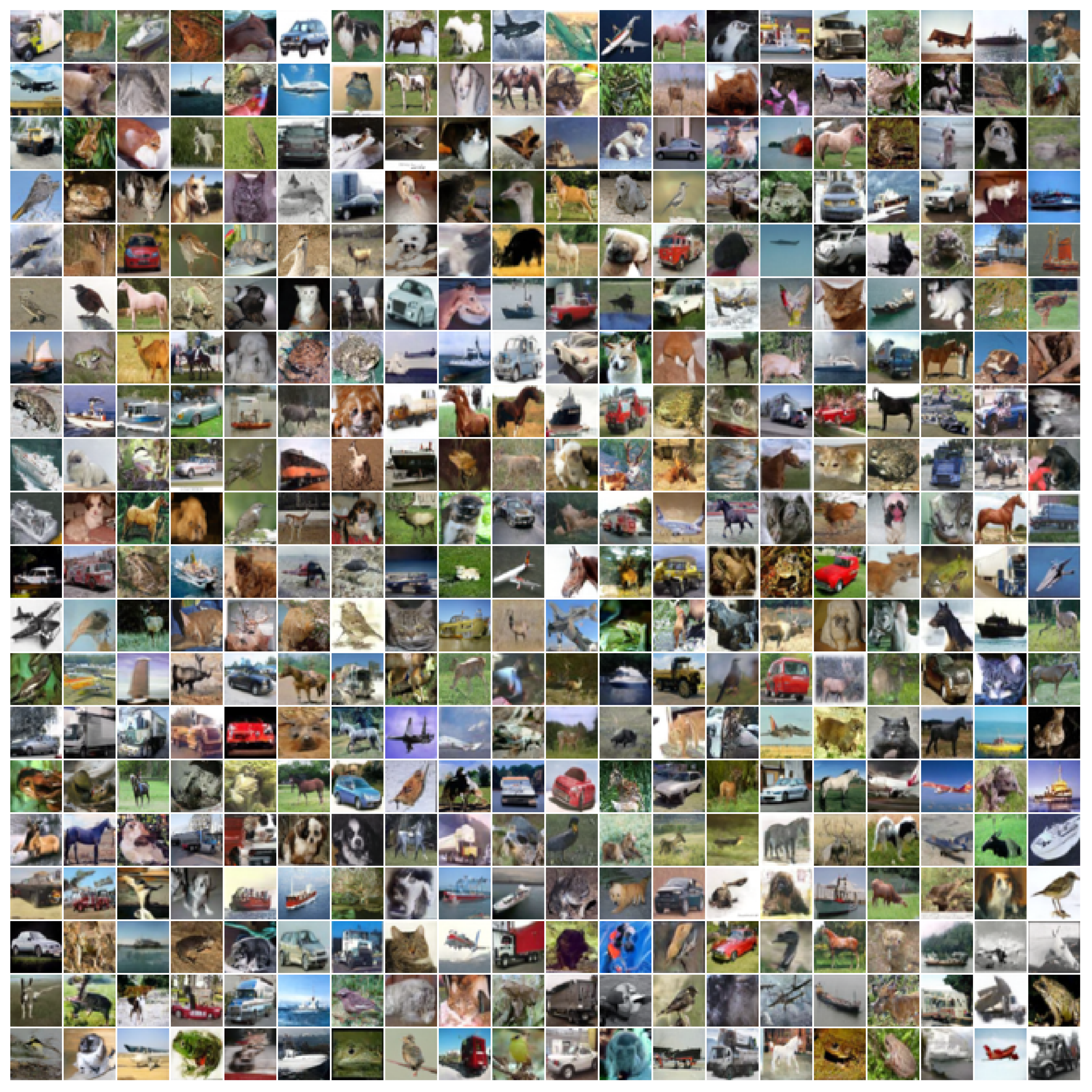}
    \caption{Four hundred samples generated by Blackout Diffusion trained by the instantaneous loss function Eq.~\eqref{eq:lossInstantaneous} and generated by $\tau$-leaping (Poisson random number) during inference.}
    \label{fig:grid-i-p}
\end{figure*}

\end{document}